\PassOptionsToPackage{table,dvipsnames}{xcolor}
\documentclass{applemlr}

\usepackage{amsmath}
\usepackage{amssymb}
\usepackage{amsfonts}
\usepackage{algorithm}
\usepackage{algpseudocode}
\usepackage{colortbl}
\usepackage{tabularx}
\usepackage{booktabs}
\usepackage{enumitem}
\usepackage{graphicx}
\usepackage{url}
\usepackage{xurl}
\usepackage{listings}
\usepackage{tikz}
\usepackage{pgfplots}
\pgfplotsset{compat=1.18}
\usetikzlibrary{arrows.meta,positioning,fit,backgrounds,calc,shapes.geometric}
\usepackage{amsmath,amsfonts,bm}

\def\eqref#1{equation~\ref{#1}}
\def\1{\bm{1}}

\DeclareMathAlphabet{\mathsfit}{\encodingdefault}{\sfdefault}{m}{sl}
\SetMathAlphabet{\mathsfit}{bold}{\encodingdefault}{\sfdefault}{bx}{n}

\newcommand{\E}{\mathbb{E}}

\newcommand{\R}{\mathbb{R}}

\makeatletter
\AtBeginDocument{\urlstyle{sf}}
\makeatother

\definecolor{aawblue}{RGB}{31,90,150}
\definecolor{aawgreen}{RGB}{34,120,80}
\definecolor{aawgray}{RGB}{90,90,90}
\definecolor{aawlight}{RGB}{235,242,250}
\definecolor{aawlightg}{RGB}{232,244,236}
\definecolor{cclight}{RGB}{253,243,238}
\definecolor{hermeslight}{RGB}{236,244,252}
\definecolor{codexlight}{RGB}{244,245,247}
\definecolor{ccorange}{RGB}{200,90,50}
\definecolor{codexgray}{RGB}{90,100,115}

\newcommand{\clate}{\textsc{SCLATE}}
\newcommand{\aaw}{\clate{}}

\newcommand{\code}[1]{\texttt{\small #1}}

\newcommand{\cmark}{\ensuremath{\checkmark}}
\newcommand{\xmark}{\ensuremath{\times}}
\newcommand{\pmark}{\ensuremath{\triangle}}

\title{\clate{}: A Substrate for Continual-Learning Agent Training and Evaluation}

\author{Youngmok Jung}
\author{Sirajul Salekin}
\author{Henry Tran}
\author{Javier Movellan}
\author{Zhao Huang}
\author{Manjot Bilkhu}

\affiliation{Apple}

\abstract{
Continual-learning agents are systems of models, harnesses, and memory operating over long multi-session horizons.
Evaluating and training them requires interleaving tasks with agent-side events such as session stop and start, crons, and memory consolidation.
Yet existing benchmarks and training frameworks schedule only the benchmark's own events, leaving each benchmark and agent pair to build a custom scheduling loop.
We present \aaw{}, an execution substrate where benchmarks and unmodified agents each add their events to one open event scheduler through an adapter.
A hybrid simulated clock runs these events on a shared timeline, flowing in real time while the agent works and skipping idle gaps, which compresses a month-long scenario into hours.
\aaw{} also serves as a rollout engine that runs any agent's harness and memory unmodified, recording the tokens and log probabilities of every model call through an in-container proxy.
We port seven benchmarks to \aaw{} and compare ten unmodified harness and memory configurations head to head on ten models.
The comparison shows that an added memory system does not reliably beat the harness's native memory and that models differ widely in how they use the same harness and memory.
We then post-train Qwen3.5-4B through unmodified harnesses and memory systems.
The model learns to use both, reading 6.8$\times$ fewer file lines with a 16.7-point higher SWE-bench Verified pass rate, and writing richer memory records, while its held-out MetaClaw accuracy rises by up to 11.8 points.
}

\metadata[Correspondence]{\sffamily Youngmok Jung: \url{yjung24@apple.com}, Manjot Bilkhu: \url{mbilkhu@apple.com}}
\date{\sffamily\today}

\begin{document}

\maketitle

\section{Introduction}

A continual-learning agent is a system comprising a memory layer that retains accumulated experience, a harness that manages tools and context, and model training that refines its behavior~\citep{wang2023voyager,park2023generative,chhikara2025mem0buildingproductionreadyai,ouyang2026reasoningbank,sun2024ttt,shenfeld2026self,lee2026meta,zhang2026agentic}.
Increasingly, agent platforms such as Claude Code, Codex, OpenClaw, and Hermes operate as such systems, learning across sequential sessions to accumulate a user's preferences, facts, habits, and workflow state.
Across such a lifecycle, the agent and its memory system initialize within an environment, execute interactive tasks while an independent grading event scores state transitions, trigger background routines to consolidate memories across session boundaries, and resume across days or weeks on a progressing timeline.
Evaluating and training these systems requires an execution substrate that runs an unmodified agent and its memory system alongside benchmark scenarios under reproducible conditions.

% EXisting evaluation and post-training substrate do not jointly support unmodified external harnesses, persistent multi-session state, and agent-side lifecycle events on a shared benchmark timeline
Yet no evaluation or training substrate can run an unmodified harness and memory across a multi-session continual-learning lifecycle.
First, non-agentic benchmarks evaluate one-shot retrieval or prompt responses on pre-collected traces, lacking temporal horizons, session boundaries, or live agent loops~\citep{wu2024longmemeval,maharana2024locomo,jiang2025know,li2026horizonbench}.
Second, closed-agent benchmarks model multi-session scenarios but bind execution to built-in assistants, testing memory or skill modules only within private runners rather than supporting external harnesses~\citep{patil2025berkeley,jiang2026amemgym,zhong2026skilllearnbench}.
Third, agent-agnostic benchmarks accept external harnesses, yet keep event scheduling closed to the agent, dictating scenario turns while preventing agents from registering crons, session stop and start, or memory consolidation onto the shared timeline~\citep{shi2026harbor,jimenez2024swebenchlanguagemodelsresolve,xia2026metaclaw,froger2026gaia2}.
On the training side, post-training rollout engines inherit these limitations by training agents only on existing single-session benchmarks, omitting session boundaries, persistent memory states, and asynchronous event simulation~\citep{xu2026polar,he2026agent,yu2026openforgerl,tan2025rllm}.
Without a shared substrate to coordinate benchmark scenarios alongside agent-side events, researchers must build custom scheduling loops for every benchmark-agent pair, preventing controlled, head-to-head comparisons.

\begin{figure}[t]
\centering
\includegraphics[width=1\linewidth]{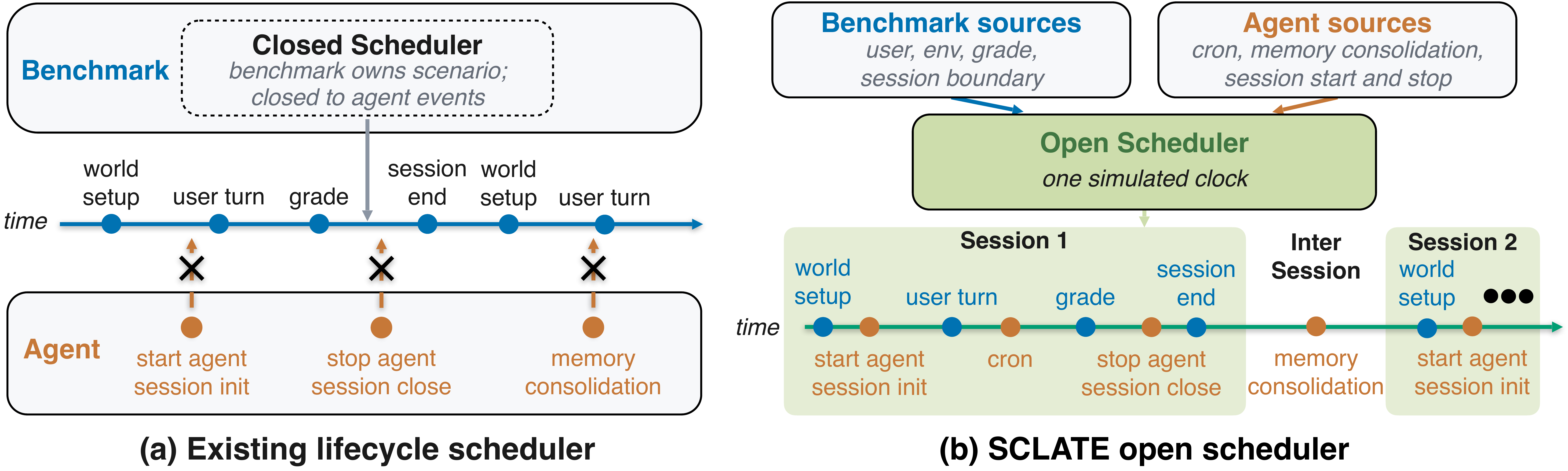}
% \vspace{-6pt}
\caption{Continual-learning lifecycle across multi-session horizons.
(a) Existing frameworks run only a built-in assistant or schedule only the benchmark's own events, leaving agent-side events off the timeline.
(b) \aaw{}'s open event scheduler lets the benchmark and the agent register their events on one simulated clock, so the same runs serve multi-session evaluation and post-training.
}
% \vspace{-6pt}
\label{fig:overview}
\end{figure}

To address these limitations, we introduce the Substrate for Continual Learning Agent Training and Evaluation (\aaw{}), an open execution substrate where arbitrary continual-learning agents can be evaluated and trained across long-horizon lifecycles spanning days to weeks (Figure~\ref{fig:overview}).
\aaw{} has three core mechanisms.
First, in a continual-learning lifecycle both the benchmark and the agent produce events. 
\aaw{} provides each with an adapter that registers those events on one open event scheduler, from scenario tasks to session stop and start, crons, and state save and restore.
Second, to run multi-week scenarios in hours, the scheduler places these events on a hybrid simulated clock, visible to agent and memory processes, that flows in real time while the agent works and jumps across idle intervals between events.
Third, to post-train models through unmodified harnesses and memory systems, an in-container proxy records every model call's tokens and log probabilities and tags each call with the harness or memory component that made it.
\aaw{} pairs this with a decoupled rollout engine that controls agent, memory, and container state across training iterations behind one training-framework-agnostic interface, which we demonstrate with Slime~\citep{zhu2025slime}.
% First, in a continual-learning lifecycle, both the benchmark and the agent act as independent event producers.
% To coordinate these independent streams, adapters decouple both benchmarks and agents under Inversion of Control, feeding an open event scheduler that interleaves benchmark scenario tasks with session stop and start, session lifecycle hooks, scheduled tasks, and state save and restore.

We validate \aaw{} on seven benchmarks, MetaClaw~\citep{xia2026metaclaw}, GAIA-2~\citep{froger2026gaia2}, AppWorld~\citep{trivedi2024appworld}, LongMemEval~\citep{wu2024longmemeval}, PersonaMem~\citep{jiang2025know}, SWE-Gym~\citep{pan2024swegym}, and SWE-bench Verified~\citep{jimenez2024swebenchlanguagemodelsresolve}, spanning personal assistance, software engineering, and multi-session memory.
With every harness and memory system running unmodified on the same benchmarks, \aaw{} enables a head-to-head comparison of continual-learning agents across three harnesses, Claude Code, Hermes, and Codex, with their native memory or an external memory system, Mem0~\citep{chhikara2025mem0buildingproductionreadyai}, Hindsight~\citep{latimer2025hindsight}, or GBrain~\citep{gbrain2026}, and ten models, including every combination on MetaClaw.
The comparison shows that no harness or memory is best for every model and that an added memory system does not reliably beat the harness's native memory.
Models also differ widely in how they use the same harness and memory, so continual-learning agents must be evaluated and trained with their model, harness, and memory together.
\aaw{} also supports post-training a model through its unmodified harness and memory. 
The model learns to use both, reading 6.8$\times$ fewer file lines with a 16.7-point higher SWE-bench Verified pass rate, and writing richer memory records, while its held-out accuracy on 30-workday MetaClaw rises by up to 11.8 points across configurations.
\aaw{} thus lets a continual-learning agent's model, harness, and memory be evaluated and trained together.

% For training, \aaw{} post-trains models through the exact same substrate, by on-policy distillation across 30-day MetaClaw lifecycles under Claude Code and Hermes and by GRPO on SWE-Gym under Claude Code and Codex, demonstrating unified evaluation and post-training on a shared execution foundation.
% For evaluation, \aaw{} runs three agent harnesses (Hermes, Codex, and Claude Code) across ten proprietary and open models (Haiku-4.5, Sonnet-5, Opus-4.6, Opus-4.8, Gemini-3.7F, GPT-5.4, GLM-5.3F, Qwen3.8-Flash-Next, DS-Vision, and DS-V4.1).

In summary, our contributions are the following.
\begin{itemize}[leftmargin=*,nosep]
\item An open event scheduler (\S\ref{sec:scheduler}) that owns the timeline and runs events from both benchmarks and unmodified agents on it, each added through an adapter (\S\ref{sec:adapters}), compressing month-long scenarios into hours while preserving event order and timing (\S\ref{sec:scheduler_results}).
\item A rollout engine (\S\ref{sec:architecture}) whose in-container capture proxy records every model call of an unmodified harness and memory, turning evaluation runs into training data.
\item \aaw{}, our implementation with seven ported benchmarks, used for a head-to-head comparison of ten harness and memory configurations on ten models (\S\ref{sec:eval_results}) and for post-training through unmodified harnesses and memory (\S\ref{sec:training_results}).
\end{itemize}

% In summary, we make four primary contributions.
% \begin{itemize}[leftmargin=*,nosep]
% \item We propose an open event scheduler that lets agents register events, such as session stop and start, crons, and memory consolidation, on a shared simulated clock alongside the benchmark's events.
% \item We build \aaw{} on this scheduler, with adapters that plug in benchmarks and unmodified harnesses and memory, and a capture proxy that turns evaluation runs into training data.
% \item We port seven benchmarks to \aaw{} and run the first head-to-head comparison of harness and memory configurations on ten models, which can be extended with new benchmarks and agents.
% \item We show that \aaw{} post-trains a model through its unmodified harness and memory across multi-session runs, and that the trained model uses both better.
% \end{itemize}

\section{Related Work} \label{sec:related}

\noindent \textbf{Continual-learning evaluation benchmarks.}
Evaluating continual-learning agents requires measuring how systems of models, harnesses, and memory persist and adapt across long multi-session horizons.
Prior evaluation suites fall into three architectural tiers.
First, non-agentic benchmarks score isolated dimensions such as memory retention and context compaction on pre-collected traces without an interactive agent loop~\citep{maharana2024locomo,wu2024longmemeval,jiang2025know,li2026horizonbench,hao2026memops}, failing to evaluate the end-to-end behavior of continual-learning agents across session boundaries.
Second, closed-agent benchmarks evaluate memory retention, skill acquisition, and model adaptation, but hardwire a built-in agent directly into the evaluation runner rather than supporting external harnesses and memory systems~\citep{patil2025berkeley,shen2026mem2act,jiang2026amemgym,zhong2026skilllearnbench}.
Third, agent-agnostic benchmarks advance to evaluating unmodified agents~\citep{xia2026metaclaw,froger2026gaia2,shi2026harbor}.
However, they either lack an event scheduler entirely~\citep{xia2026metaclaw} or rely on benchmark-driven execution cycles that treat the agent as a passive responder~\citep{froger2026gaia2,shi2026harbor}, leaving no mechanism to coordinate agent and memory events across continual-learning lifecycles.
\aaw{} decouples the event scheduling control loop from the benchmark into an open scheduler, supporting the broader continual-learning lifecycle events of agents and benchmarks.

\noindent \textbf{Rollout engines for agent post-training.}
A growing line of systems scales agent post-training by decoupling rollout execution from distributed training loops~\citep{he2026agent,tan2025rllm,xu2026polar,zhu2025slime,cao2025skyrl,primeintellect2026prime,zhang2026prorl}.
To train unmodified agents, recent frameworks intercept model calls via network proxies~\citep{xu2026polar,he2026agent,yu2026openforgerl} or wrap execution loops in standardized abstractions~\citep{tan2025rllm}.
However, training continual-learning agents requires carrying session state across container restarts and attributing each model call to the harness or memory component that made it.
\aaw{} addresses these requirements by providing abstractions to manage rollout state across container lifecycles and the flexibility to route and attribute model calls by component, natively supporting continual-learning agents at scale while remaining agnostic to benchmarks and training frameworks.
Table~\ref{tab:related-capabilities} in Appendix~\ref{app:related_comparison} compares \aaw{} with these systems.
% A continual-learning agent consists of a model, a harness, and a memory system, and the harness and memory system generate their own agent-side events, such as session start and shutdown, crons, and memory consolidation. 
% Running such an agent as-is requires a timeline on which these events execute faithfully alongside benchmark events, and an open event contract through which the substrate can interpret them. None of the existing evaluation or post-training systems runs agent-side events alongside the benchmark, and \aaw{} provides both the shared timeline and the open event contract.

\section{The \aaw{} Substrate} \label{sec:framework}

\begin{figure*}[t]
\centering
\includegraphics[width=\linewidth]{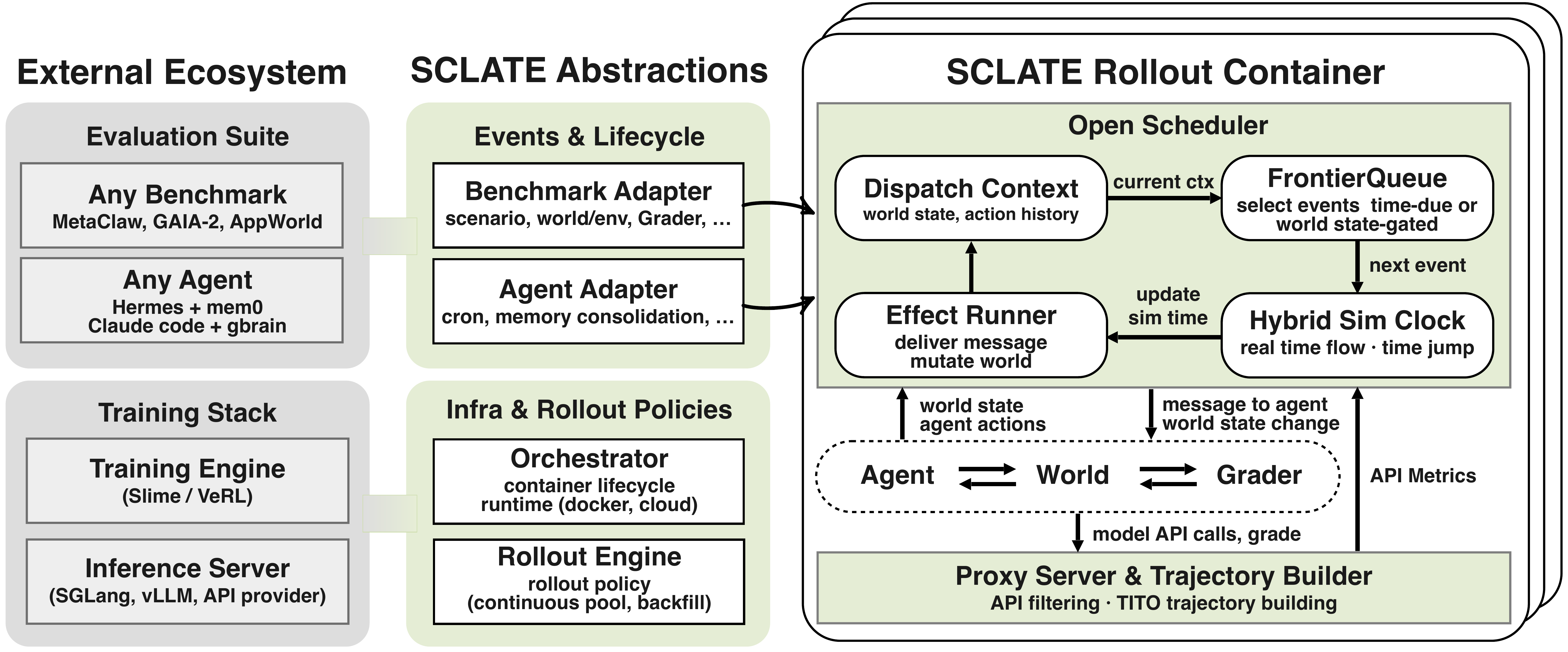}
\vspace{-18pt}
\caption{\aaw{} architecture.
Adapters decouple benchmarks and agent harnesses into a uniform event stream, while the host orchestration manages container lifecycles and training batches.
Inside replicated sandboxes, an open event scheduler coordinates benchmark and agent events on a hybrid simulated clock, while an in-container proxy captures token-faithful trajectories for post-training.
}
\vspace{-13pt}
\label{fig:framework}
\end{figure*}

\noindent \textbf{Architecture overview.}
\aaw{} is an execution substrate for continual-learning agent evaluation and post-training across multi-session horizons (Figure~\ref{fig:framework}).
\aaw{} is organized into three decoupled tiers, an external tier of benchmarks and agent harnesses, a host orchestration plane that manages container lifecycles and training batches, and replicated container sandboxes that execute scenarios in isolation.
The system provides standardized abstractions for event scheduling, agent lifecycles, and sandboxed execution, allowing new benchmarks and external agent harnesses to plug in without modifying substrate code.
At its core, an open event scheduler coordinates scenario progression, agent-side routines, and task grading on a shared hybrid simulated clock.
An in-container capture proxy, through which all model calls pass, pairs live evaluation with on-policy trajectory capture.
Throughout the paper, an unmodified harness is run without changes to its source code and integrates only through an adapter, configuration files such as hook settings, and environment variables such as the model endpoint pointed at the capture proxy and the libfaketime preload.

\subsection{Event Scheduling Loop and Hybrid Clock} \label{sec:scheduler}
A continual-learning lifecycle interleaves benchmark events (scenario tasks, environment mutations, session boundaries) with agent-side routines (process bootstrap, background crons, memory consolidation, and state handoffs).
Running it in wall-clock time is prohibitively slow, while naive discrete-event simulation breaks execution by advancing time while an agent is still executing tools.
\aaw{} avoids both with an open event scheduler to which benchmark adapters and agent adapters register event sources, coordinated over a hybrid simulated clock.

\noindent \textbf{Event scheduling loop.}
At each tick, the scheduler snapshots simulated time, world state, agent actions, and grading outcomes into a DispatchContext.
The FrontierQueue then selects the registered event sources that are due, by time or by a condition on the world state, agent actions, or grading outcomes.
The simulated time controller advances the clock, and the Effect Runner applies the chosen event, delivering a prompt or notification, mutating the world, or starting, stopping, saving, or restoring the agent and its memory.
Appendix~\ref{sec:appendix_interfaces} gives the full loop (Algorithm~\ref{alg:sclate_scheduling_loop}) and the catalogs of event sources and effect primitives (Tables~\ref{tab:event_sources} and~\ref{tab:effects}).

\noindent \textbf{Hybrid simulated clock.}
The clock flows with real time while the agent is busy or a background job holds the clock, so that tools, test suites, and internal timers run naturally.
Once nothing is busy, the future event stream is fixed, and the clock jumps across the idle interval to the earliest scheduled event, compressing overnight gaps and session boundaries into pure execution time, as traced on one MetaClaw workday in Figure~\ref{fig:walk-metaclaw} (Appendix~\ref{app:scheduler_walkthroughs}) and measured in Table~\ref{tab:scheduler-metrics}.
libfaketime~\citep{libfaketime_correct} makes standard system-time calls return the simulated time for every dynamically linked harness and memory process evaluated here, so native crons in Claude Code and Hermes follow the scenario timeline without source changes (Figure~\ref{fig:time-grounding-evidence} in Appendix~\ref{app:time_grounding}).

\subsection{Pluggable Benchmark and Agent Adapters} \label{sec:adapters}
\aaw{} coordinates the benchmark and agent lifecycles through two adapters, the Benchmark Adapter and the Agent Adapter, both of which register event sources with the scheduler (Table~\ref{tab:interfaces} in Appendix~\ref{sec:appendix_interfaces}).
The adapters translate scenario progression and agent routines into the shared event language, enabling controlled, head-to-head evaluation of unmodified harnesses across benchmarks.

\noindent \textbf{Benchmark adapter and scenario composition.}
A benchmark needs its world state set up, benchmark events delivered, and turns graded.
The Benchmark Adapter plugs a benchmark into \aaw{} through four components (Table~\ref{tab:interfaces}).
Its setup routine establishes the baseline world state at container start, configuring code repositories, seeding databases, and starting benchmark application servers.
Its source manifest turns the benchmark events (scenario tasks, environment mutations, and session boundaries) and the graded turns into \aaw{}'s event sources.
An optional world provider exposes a read-only view of application state to event sources and graders through the DispatchContext.
A grading policy declares which turns are graded and when, and an independent grader scores them against the task rubric, so the scheduler carries no benchmark-specific grading logic.

\noindent \textbf{Agent adapter and state persistence.}
Continual-learning agents evolve across multi-session lifecycles where memory systems persist durable state, while the agent process stops and starts at session boundaries.
The Agent Adapter connects the open scheduler to an unmodified harness and its memory system through interfaces for the session lifecycle, message delivery, and state save and restore, and registers the agent's own event sources with the scheduler (Table~\ref{tab:interfaces}).
During container startup and session transitions, the substrate polls the adapter's readiness probe until the memory system and the agent process can receive turns, the adapter delivers each message the scheduler dispatches into the live agent binary without blocking and parses the agent's response from it, and the adapter ends each session at its boundary, where hooks attached to session-start and session-end events can flush session state, persist memory, or run custom scripts.
The adapter also registers the agent's own event sources, such as session-start and handoff prompts, background crons, and memory consolidation, while crons and wakeups that a harness schedules internally run on the simulated clock and keep it flowing.
% Registering events through the adapter lets a memory system's maintenance schedule be prototyped without changing its code.
The scheduler performs state persistence across sessions by consulting the adapter's memory specification, which declares the agent and the memory-system state separately and defines how to flush, snapshot, and restore agent files and sidecar databases.
% Save and restore take a scope argument that selects the agent, the memory system, or both, and write either a directory snapshot or a sidecar export.

\vspace{-5pt}
\subsection{Host Orchestration and Rollout Infrastructure} \label{sec:architecture} \label{sec:training}

Scaling multi-session agent evaluation and post-training rollouts requires coordinating parallel execution across heterogeneous infrastructure while isolating complex benchmark environments.
Standardized container environments initialize reproducible workspaces across parallel instances while preventing evaluation leakage.
The host control plane organizes execution through the Orchestrator, which manages scenario configurations and container lifecycles, and the Rollout Engine, which coordinates distributed rollout collection for post-training.
This division of responsibilities keeps benchmark dependencies, agent execution, and trajectory capture self-contained across parallel runs.

\noindent \textbf{Pluggable runtime and rollout abstractions.}
The Orchestrator manages container lifecycles through a unified Runtime protocol that presents the same execution semantics across local and distributed infrastructure.
This protocol defines typed interfaces for container provisioning, job submission, non-blocking health polling, and artifact retrieval, with adapters supporting local Docker engines or a managed cloud cluster.
Building on this runtime foundation, the Rollout Engine provides abstractions to manage stateful container lifecycles across multi-session horizons.
Leveraging the adapter's memory specification, the engine exposes the adapter's save and restore primitives, selected by scope, to move agent and memory state across container instances, while leaving task selection and training steps to the external trainer stack.
This standardized interface abstracts parallel stateful execution into structured trajectory batches, interfacing with external post-training frameworks, such as reinforcement learning and on-policy distillation, without coupling lifecycle management to specific training algorithms.
\aaw{} validates this complete host abstraction across local Docker engines and a managed cloud cluster for runtime execution, coupling the Rollout Engine with Slime~\citep{zhu2025slime} as an example distributed training stack.

\noindent \textbf{In-container capture proxy and trajectory assembly.}
To support post-training without modifying harness code, \aaw{} deploys a capture proxy server directly inside each rollout container.
The capture proxy operates as the dedicated model API gateway inside the container, recording prompt tokens, completions, and log probabilities (example in Figure~\ref{fig:rollout-tokens}) while routing requests to local inference engines or cloud providers based on model type and API keys.
Through configurable tagging rules over API keys, system prompt signatures, and tool schemas, the proxy categorizes each invocation, allowing downstream trainer adapters to distinguish the agent's own turns from auxiliary calls, such as the web-search calls excluded from SWE-Gym training.
The proxy assembles the recorded calls into token-level trajectories, and on the host the Rollout Engine launches rollouts and gathers these trajectories with their log probabilities and grading verdicts into training batches for external post-training frameworks.
% Because every model call passes through it, the proxy also feeds real-time activity metrics into the DispatchContext, so the scheduler detects background execution such as self-wakeups and internal cron routines that run without external message delivery.

\noindent \textbf{In-container privilege separation.}
\aaw{} keeps evaluation data out of the agent's reach by running the agent under a separate unprivileged user, so the agent cannot read ground truth or grading hints.
The daemon user runs the event scheduler and benchmark services and owns the scenario ground truth, run artifacts, and grading and scheduling code, in directories the agent user cannot list or read.
The agent user runs the harness and its shell tools in its own workspace and reaches the daemon only through message channel files that agent and daemon users can read and write.

\begin{figure*}[t]
\centering
\includegraphics[width=\linewidth]{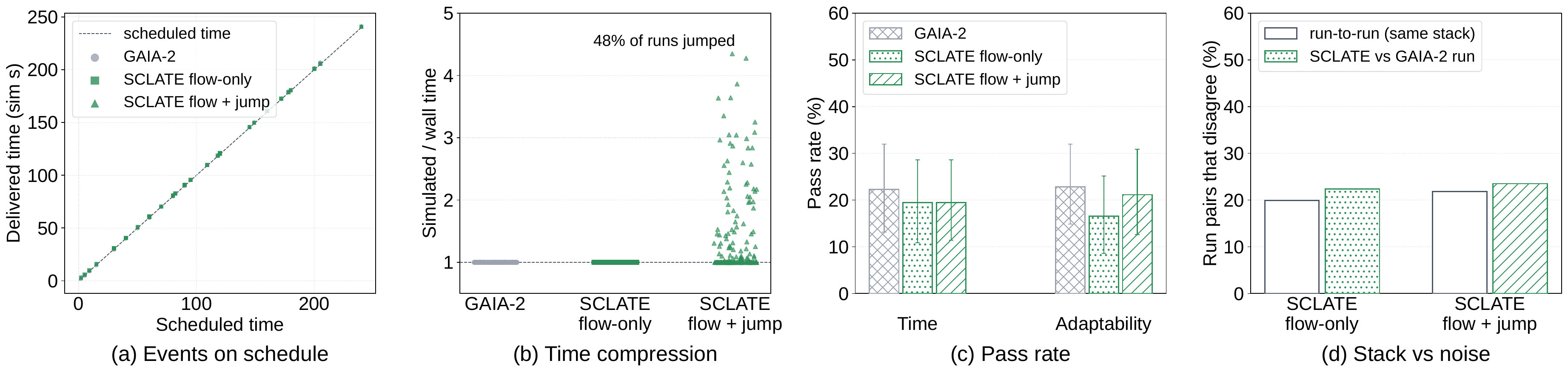}
\vspace{-17pt}
\caption{
Parity between the upstream Gaia2-CLI stack and \aaw{}, with the same agent and judge.
(a) Every scheduled environment event is delivered on time on all three stacks.
(b) On the Time scenarios, idle-time jumps compress simulated time by up to 4.3$\times$; the label gives the share of runs that clock jumped.
(c) Pass rate on 25 Time and 25 Adaptability scenarios with Haiku-4.5, each scenario repeated 7 times on every stack (175 runs per bar, 1,050 runs in total).
(d) An \aaw{} run and a GAIA-2 run of the same scenario disagree about as often as two runs of the same stack.
}
\vspace{-10pt}
\label{fig:gaia2-parity}
\end{figure*}

\begin{table*}[t]
\caption{
  MetaClaw pass accuracy over 30-workday arc (346 rounds) across ten models and ten harness and memory configurations.
  % Native refers to each harness's built-in memory mechanism (Claude Code auto-memory, Hermes skills with notes, and Codex native memory).
  % Mem0, Hindsight, and GBrain run on top of each harness's native memory and make their own extraction and consolidation calls with the same model as the agent, so each of their columns ablates the added memory system against the Native column.
  % Cell color intensity scales continuously with accuracy within each harness family.
  Bold marks the best accuracy within each harness-model pair.
}
\label{tab:metaclaw-matrix}
\centering
\scriptsize
\setlength{\tabcolsep}{2.5pt}
\renewcommand{\arraystretch}{1.20}
\begin{tabular*}{\linewidth}{@{\extracolsep{\fill}}lcccccccccc@{}}
\toprule
& \multicolumn{5}{c}{\footnotesize\textbf{Claude Code}} & \multicolumn{4}{c}{\footnotesize\textbf{Hermes}} & \multicolumn{1}{c}{\footnotesize\textbf{Codex}} \\
\cmidrule(lr){2-6} \cmidrule(lr){7-10} \cmidrule(lr){11-11}
\textbf{Model} & No-Mem & Native & Mem0 & Hindsight & GBrain & No-Mem & Native & Mem0 & Hindsight & Native \\
\midrule
Haiku-4.5 & \cellcolor{ccorange!6} 40.2\% & \cellcolor{ccorange!10} 46.0\% & \cellcolor{ccorange!27} \textbf{69.7\%} & \cellcolor{ccorange!19} 58.7\% & \cellcolor{ccorange!18} 56.7\% & \cellcolor{aawblue!9} 43.6\% & \cellcolor{aawblue!20} 59.8\% & \cellcolor{aawblue!22} \textbf{61.9\%} & \cellcolor{aawblue!21} 61.0\% & \cellcolor{codexgray!9} 44.8\% \\
Sonnet-5 & \cellcolor{ccorange!9} 43.9\% & \cellcolor{ccorange!14} 51.2\% & \cellcolor{ccorange!18} 57.5\% & \cellcolor{ccorange!14} 50.9\% & \cellcolor{ccorange!22} \textbf{62.1\%} & \cellcolor{aawblue!8} 43.1\% & \cellcolor{aawblue!11} 47.4\% & \cellcolor{aawblue!20} \textbf{59.0\%} & \cellcolor{aawblue!13} 50.3\% & \cellcolor{codexgray!18} 56.7\% \\
Opus-4.6 & \cellcolor{ccorange!12} 48.8\% & \cellcolor{ccorange!24} 65.0\% & \cellcolor{ccorange!25} 66.5\% & \cellcolor{ccorange!28} 71.1\% & \cellcolor{ccorange!30} \textbf{74.0\%} & \cellcolor{aawblue!11} 46.5\% & \cellcolor{aawblue!26} \textbf{68.5\%} & \cellcolor{aawblue!24} 65.0\% & \cellcolor{aawblue!25} 66.8\% & \cellcolor{codexgray!14} 51.5\% \\
Opus-4.8 & \cellcolor{ccorange!12} 48.6\% & \cellcolor{ccorange!21} 60.4\% & \cellcolor{ccorange!25} 66.8\% & \cellcolor{ccorange!30} \textbf{74.0\%} & \cellcolor{ccorange!26} 67.6\% & \cellcolor{aawblue!10} 45.4\% & \cellcolor{aawblue!23} 64.2\% & \cellcolor{aawblue!24} \textbf{64.7\%} & \cellcolor{aawblue!22} 61.9\% & \cellcolor{codexgray!12} 48.6\% \\
Gemini-3.7F & \cellcolor{ccorange!23} 63.9\% & \cellcolor{ccorange!34} 79.2\% & \cellcolor{ccorange!34} 79.2\% & \cellcolor{ccorange!36} \textbf{81.8\%} & \cellcolor{ccorange!34} 79.5\% & \cellcolor{aawblue!30} 73.4\% & \cellcolor{aawblue!33} 77.8\% & \cellcolor{aawblue!34} 79.5\% & \cellcolor{aawblue!35} \textbf{80.1\%} & \cellcolor{codexgray!30} 73.4\% \\
GPT-5.4 & \cellcolor{ccorange!6} 40.5\% & \cellcolor{ccorange!13} 50.3\% & \cellcolor{ccorange!18} \textbf{56.4\%} & \cellcolor{ccorange!15} 53.2\% & \cellcolor{ccorange!16} 53.8\% & \cellcolor{aawblue!7} 41.0\% & \cellcolor{aawblue!16} 53.8\% & \cellcolor{aawblue!27} \textbf{68.8\%} & \cellcolor{aawblue!19} 57.8\% & \cellcolor{codexgray!7} 40.8\% \\
GLM-5.3F & \cellcolor{ccorange!11} 46.8\% & \cellcolor{ccorange!24} 65.0\% & \cellcolor{ccorange!25} \textbf{66.8\%} & \cellcolor{ccorange!25} 65.9\% & \cellcolor{ccorange!25} 66.2\% & \cellcolor{aawblue!11} 47.1\% & \cellcolor{aawblue!21} 60.4\% & \cellcolor{aawblue!26} \textbf{67.3\%} & \cellcolor{aawblue!24} 64.5\% & \cellcolor{codexgray!21} 60.4\% \\
Qwen3.8-FN & \cellcolor{ccorange!13} 49.7\% & \cellcolor{ccorange!27} 69.7\% & \cellcolor{ccorange!27} 69.4\% & \cellcolor{ccorange!25} 66.8\% & \cellcolor{ccorange!30} \textbf{74.3\%} & \cellcolor{aawblue!15} 52.9\% & \cellcolor{aawblue!28} \textbf{71.4\%} & \cellcolor{aawblue!23} 63.3\% & \cellcolor{aawblue!26} 68.2\% & \cellcolor{codexgray!28} 70.8\% \\
DS-Vision & \cellcolor{ccorange!10} 45.4\% & \cellcolor{ccorange!16} 54.6\% & \cellcolor{ccorange!25} \textbf{66.8\%} & \cellcolor{ccorange!22} 63.0\% & \cellcolor{ccorange!20} 59.0\% & \cellcolor{aawblue!13} 50.0\% & \cellcolor{aawblue!27} \textbf{69.1\%} & \cellcolor{aawblue!22} 63.0\% & \cellcolor{aawblue!22} 62.1\% & \cellcolor{codexgray!16} 53.5\% \\
DS-V4.1 & \cellcolor{ccorange!15} 52.0\% & \cellcolor{ccorange!23} 64.2\% & \cellcolor{ccorange!30} 74.3\% & \cellcolor{ccorange!30} 74.3\% & \cellcolor{ccorange!31} \textbf{74.9\%} & \cellcolor{aawblue!20} 59.3\% & \cellcolor{aawblue!30} 74.0\% & \cellcolor{aawblue!31} \textbf{74.6\%} & \cellcolor{aawblue!27} 68.8\% & \cellcolor{codexgray!30} 73.7\% \\
\bottomrule
\end{tabular*}
\vspace{-8pt}
\end{table*}

\vspace{-5pt}
\section{Experiments} \label{sec:experiments}

We now evaluate whether \aaw{} achieves our design goals.
\begin{itemize}[leftmargin=*,nosep]
\item Can \aaw{}'s open scheduler run asynchronous benchmark scenarios and agent-side events on one simulated clock while compressing idle time and preserving event order (Section~\ref{sec:scheduler_results})?
\item Can \aaw{} compare harness and memory configurations head to head across models, and what does the comparison reveal about how models use their harness and memory (Section~\ref{sec:eval_results})?
\item Can \aaw{} post-train models through unmodified harnesses and memory systems using the same runs it evaluates, including across multi-session lifecycles (Section~\ref{sec:training_results})?
\end{itemize}

\noindent \textbf{Experimental setup.}
We evaluate \aaw{} across seven ported public benchmarks spanning personal assistance, software engineering, and multi-session memory.
Detailed workload porting, dataset splits, and environment specifications are in Appendix~\ref{sec:benchmark_specs}.

\textbf{Models.} We use ten proprietary and open models: Haiku-4.5, Sonnet-5, Opus-4.6, Opus-4.8, Gemini-3.7F, GPT-5.4, GLM-5.3F, Qwen3.8-Flash-Next (Qwen3.8-FN in tables), DeepSeek-Flash-Vision, and DeepSeek-V4.1-Flash. 
Models are queried through API provider endpoints or self-hosted serving endpoints, using the same reasoning effort for a model under every harness (Table~\ref{tab:reasoning-effort} in Appendix~\ref{sec:benchmark_specs}).

\textbf{Harnesses and memory systems.} 
We evaluate three agent harnesses (Claude Code, Hermes, and Codex) paired with three memory backends (Mem0~\citep{chhikara2025mem0buildingproductionreadyai}, Hindsight~\citep{latimer2025hindsight}, and GBrain~\citep{gbrain2026}).
Native refers to each harness's built-in memory (Claude Code auto-memory and Hermes skills with notes), and No-Mem turns it off while keeping the harness tools, including session search over prior transcripts.
Mem0, Hindsight, and GBrain run on top of native memory and make their own extraction and consolidation calls with the same model as the agent.
Every harness introduces session stop and start events. Mem0 and Hindsight retain and recall memories during task execution, while GBrain registers memory consolidation events between sessions.

\textbf{Workloads.} Our ported benchmark suite includes MetaClaw~\citep{xia2026metaclaw} (30-workdays spanning 40 calendar days, 346 interactive rounds), GAIA-2~\citep{froger2026gaia2}, AppWorld~\citep{trivedi2024appworld}, two converted recall suites (LongMemEval~\citep{wu2024longmemeval}, PersonaMem~\citep{jiang2025know}), SWE-Gym~\citep{pan2024swegym}, and SWE-bench Verified~\citep{jimenez2024swebenchlanguagemodelsresolve}.

\begin{figure}[t]
\centering
\includegraphics[width=\linewidth]{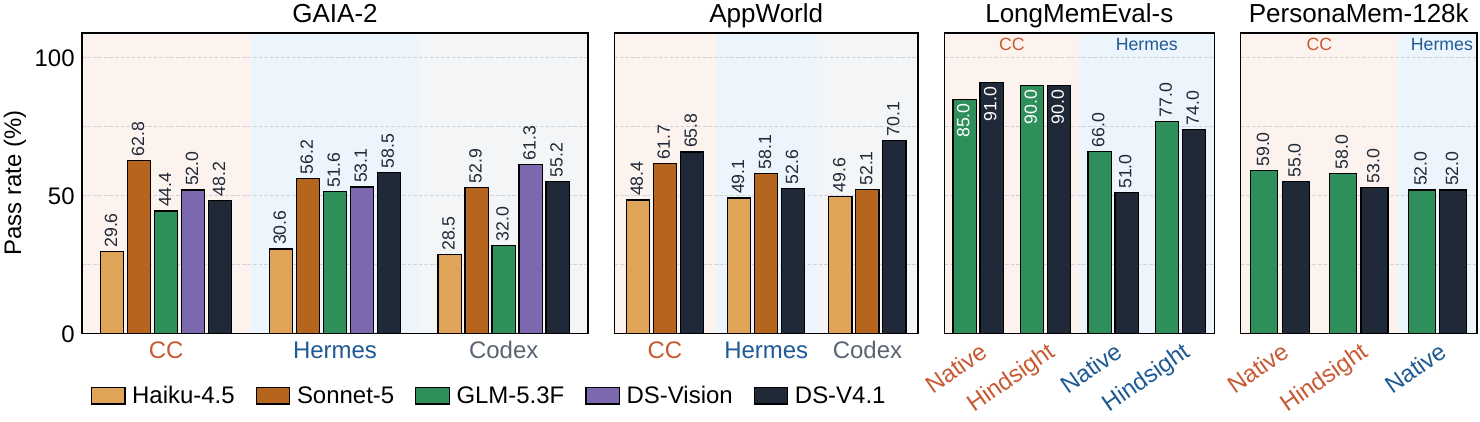}
\vspace{-23pt}
\caption{Pass rate (\%) of each model under each harness and memory configuration; CC is Claude Code.
\textbf{Left: GAIA-2 and AppWorld,} with each harness on its native memory. No harness is best for every model.
\textbf{Right: LongMemEval-s and PersonaMem-128k,} with Claude Code and Hermes on native memory or an added memory system. Claude Code recalls more than Hermes on LongMemEval-s, and the two are close on PersonaMem-128k.
Hermes with Hindsight is not shown on PersonaMem-128k, where its per-turn memory updates exceed the 900\,s turn limit.
}
\vspace{-15pt}
\label{fig:results-overview}
\end{figure}

\subsection{Validating the Open Event Scheduler} \label{sec:scheduler_results}

We validate \aaw{}'s open event scheduler against upstream GAIA-2 by porting its scenarios as an ordinary benchmark adapter.
GAIA-2 schedules its events asynchronously, so they can arrive while the agent is working, and triggers them by time or by world state, such as a reply that follows the agent's own message after a set delay~\citep{froger2026gaia2}.
Its scheduler is closed, owning the clock and advancing it in real time or accelerated only when the agent calls a wait tool, so only the scenario can schedule events.
We run the same Hermes agent with Haiku-4.5 and the same judge on 25 Time and 25 Adaptability scenarios, on upstream GAIA-2 and on \aaw{} with its clock either only flowing or flowing and jumping (Figure~\ref{fig:gaia2-parity}).
Every scheduled environment event is delivered on all three stacks, in the same order and within 0.9\,s of its scheduled time at the median and 2.0\,s at most (Figure~\ref{fig:gaia2-parity}a).
The three stacks reach overlapping pass rates (Figure~\ref{fig:gaia2-parity}c, with the grading in Appendix~\ref{app:gaia2_parity}), and an \aaw{} run and a GAIA-2 run of the same scenario disagree about as often as two runs of the same stack (Figure~\ref{fig:gaia2-parity}d), so the port changes outcomes no more than rerunning the agent does.
Across all evaluation, the scheduler fires all 13{,}050 GAIA-2 environment events, and 98.7\% of the checkable ones at the delay the scenario specifies (Table~\ref{tab:scheduler-metrics}, with walkthroughs in Figures~\ref{fig:walk-gaia2-adapt}--\ref{fig:walk-gaia2-midturn}).

We then measure how far the hybrid clock compresses scenarios and check whether the harness and memory run on simulated time.
On GAIA-2 Time scenarios the clock jumps in about half the runs, compressing a run by up to $4.3\times$ (Figure~\ref{fig:gaia2-parity}b), and across all 100 MetaClaw runs of Table~\ref{tab:metaclaw-matrix} a 30-workday arc spanning 40 calendar days finishes in a median of 5.9 real hours, a median speed-up of $162\times$ ($142\times$ over all runs combined) (Table~\ref{tab:scheduler-metrics} in Appendix~\ref{app:scheduler_metrics}).
Under Claude Code, Codex, and Hermes, the simulated date reaches the harness's system prompt, its shell clock queries, and the timestamps that Mem0 and Hindsight record (Figure~\ref{fig:time-grounding-evidence}), and harness crons and GBrain's nightly consolidation fire on the simulated timeline.
Appendix~\ref{app:scheduler_walkthroughs} traces the mechanism on individual GAIA-2 and MetaClaw runs.

\begin{table}[t]
\centering
\scriptsize
\renewcommand{\arraystretch}{1.20}
\setlength{\tabcolsep}{3.5pt}
\begin{tabular}{lcccccccc}
\toprule
& \multicolumn{4}{c}{\textbf{In-distribution (days 1--20)}} & \multicolumn{4}{c}{\textbf{Held-out (days 21--30)}} \\
\cmidrule(lr){2-5} \cmidrule(lr){6-9}
\textbf{Harness \& Memory} & Base & Trained & $\Delta$ (pp) & $p$ & Base & Trained & $\Delta$ (pp) & $p$ \\
\midrule
CC \& no-memory & 37.4\% & 52.4\% & $+15.0$          & $<$0.001 &  9.2\% & 16.8\% & $+7.6$           & 0.039 \\
CC \& Mem0      & 42.7\% & 67.4\% & $\mathbf{+24.7}$ & $<$0.001 & 10.1\% & 21.8\% & $\mathbf{+11.8}$ & 0.023 \\
CC \& Hindsight & 51.1\% & 52.0\% & $+0.9$           & 0.91     & 16.0\% & 16.8\% & $+0.8$           & 0.83 \\
\midrule
Hermes \& no-memory & 29.1\% & 36.6\% & $+7.5$           & 0.27     &  8.4\% & 18.5\% & $\mathbf{+10.1}$ & 0.047 \\
Hermes \& Mem0      & 31.3\% & 58.6\% & $\mathbf{+27.3}$ & 0.004    &  8.4\% & 16.0\% & $+7.6$           & 0.17 \\
Hermes \& Hindsight & 39.6\% & 58.6\% & $+18.9$          & $<$0.001 & 10.1\% & 20.2\% & $\mathbf{+10.1}$ & 0.051 \\
\bottomrule
\end{tabular}
\caption{MetaClaw post-training on a Qwen3.5-4B model across six configurations.
Columns show base and post-trained (epoch 5) accuracy and the improvement ($\Delta$) on the in-distribution and held-out partitions.
$p$ is a paired wild cluster bootstrap-$t$ test with simulated days as clusters.}
\label{tab:sclate-training-metaclaw}
\vspace{-8pt}
\end{table}

\subsection{Evaluating Continual-Learning Agents} \label{sec:eval_results}

\noindent \textbf{Continual-learning evaluation on MetaClaw.}
Upstream MetaClaw~\citep{xia2026metaclaw} runs a fixed memory system and simulates time through dates injected into prompts.
\aaw{}'s MetaClaw adapter replaces those dates with the simulated clock, so unmodified harnesses (Claude Code, Hermes, Codex) and memory systems (Mem0, Hindsight, GBrain) run together across 30 simulated workdays.

Table~\ref{tab:metaclaw-matrix} reports the full grid of ten harness and memory configurations across ten models over 346 rounds.
Every memory configuration outperforms the no-memory baseline of its harness for all models, but no single memory system is best for every model under either harness.
Against native memory alone, an added memory system scores lower in 11 of 50 cells, 8 of them under Hermes, whose native skills and notes are strong.
On Haiku-4.5, harness workflows shape task success, lifting Hermes Native to 59.8\% versus 46.0\% for Claude Code, which trajectory inspection attributes to dedicated skill-management tools and a hook that requires verification after tool calls.
On most other models this harness gap narrows or reverses, with Claude Code Native matching Hermes Native within 1.4 percentage points (pp) on Gemini-3.7F and leading it on Sonnet-5, though DS-Vision shows a similar 14.5 pp gap.
Gemini-3.7F also uses the harness's session-search tool to inspect prior session transcripts, spending 7.6\% of its tool calls (2{,}254 of 29{,}550) on session search versus at most 1.6\% for other models, and is the only model to search in all ten configurations (Figure~\ref{fig:metaclaw-session-search} in Appendix~\ref{app:evaluation_details}).
This behavior recovers procedural rules from earlier sessions without any memory system, consistent with its lead in no-memory accuracy, while the other models rarely call the same tool.
These results show that an added memory system does not reliably beat the harness's native memory and that models differ widely in how they use the same harness and memory.

\noindent \textbf{GAIA-2 and AppWorld results.}
On the 800 GAIA-2 scenarios (Figure~\ref{fig:results-overview}, per category in Table~\ref{tab:gaia2-full-breakdown}), Search and Execution are the easiest categories in all 15 harness--model cells, as in the original release, while Adaptability and Time, whose scenarios depend on events delivered during the run, stay low in every cell (at most 37.5\% and 34.4\%).
The strongest model differs by harness, with Sonnet-5 leading under Claude Code (62.8\%), DS-V4.1 under Hermes (58.5\%), and DS-Vision under Codex (61.3\%), and because generation consumes simulated time as in GAIA-2's default time mode, these scores also reflect each harness's and endpoint's latency.
On AppWorld, harnesses reach 48.4\%--49.6\% on Haiku-4.5, 52.1\%--61.7\% on Sonnet-5, and up to 70.1\% on DS-V4.1 (Figure~\ref{fig:results-overview}).

\noindent \textbf{Multi-session recall benchmarks.}
We convert two memory suites, LongMemEval and PersonaMem, into multi-session benchmarks.
Their upstream formulations score retrieval over pre-collected conversation logs in a single static prompt, with no process boundaries, background maintenance routines, or live tool use.
Our adapters split each conversation history into sessions on consecutive simulated days, stop the agent process between sessions, and ask the question in the last session, so the agent must carry earlier facts through its native memory or a memory system, while the dataset content, questions, and ground-truth answers stay unchanged.
On LongMemEval-s, Claude Code's native memory outscores Hermes's, 85 and 91 against 66 and 51 on GLM-5.3F and DS-V4.1 (Figure~\ref{fig:results-overview}).
Hermes's built-in memory has a small size cap and rejects new facts once full, so most facts are never stored, and adding Hindsight, which has no such cap, lifts Hermes's accuracy to 77 and 74 on GLM-5.3F and DS-V4.1.
% TODO(recall results): one sentence of headline results, e.g. "Across ..., ... (Table~\ref{tab:recall-results} in Appendix~\ref{app:recall_details})."

\subsection{Post-Training Continual-Learning Agents} \label{sec:training_results}

\begin{figure*}[t]
\centering
\includegraphics[width=0.7\linewidth]{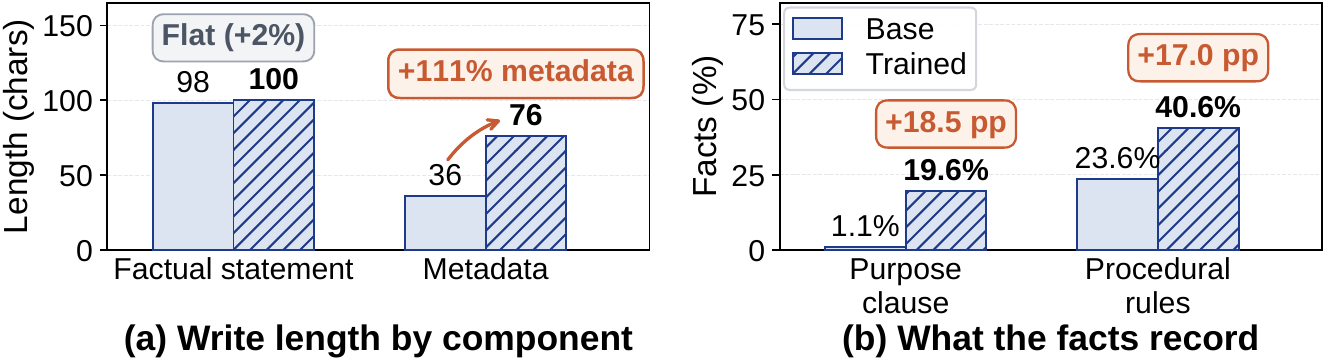}
\vspace{-6pt}
\caption{Memory facts written by the base and trained models as Hindsight's fact extractor on Hermes (5{,}027 and 3{,}617 facts).
(a) Median length: training lengthens the metadata (+111\%) but not the statement (+2\%).
(b) Facts containing the optional purpose clause or a procedural rule.
}
\label{fig:memwrite-isolation}
\vspace{-4pt}
\end{figure*}

\begin{figure*}[t]
\centering
\includegraphics[width=\linewidth]{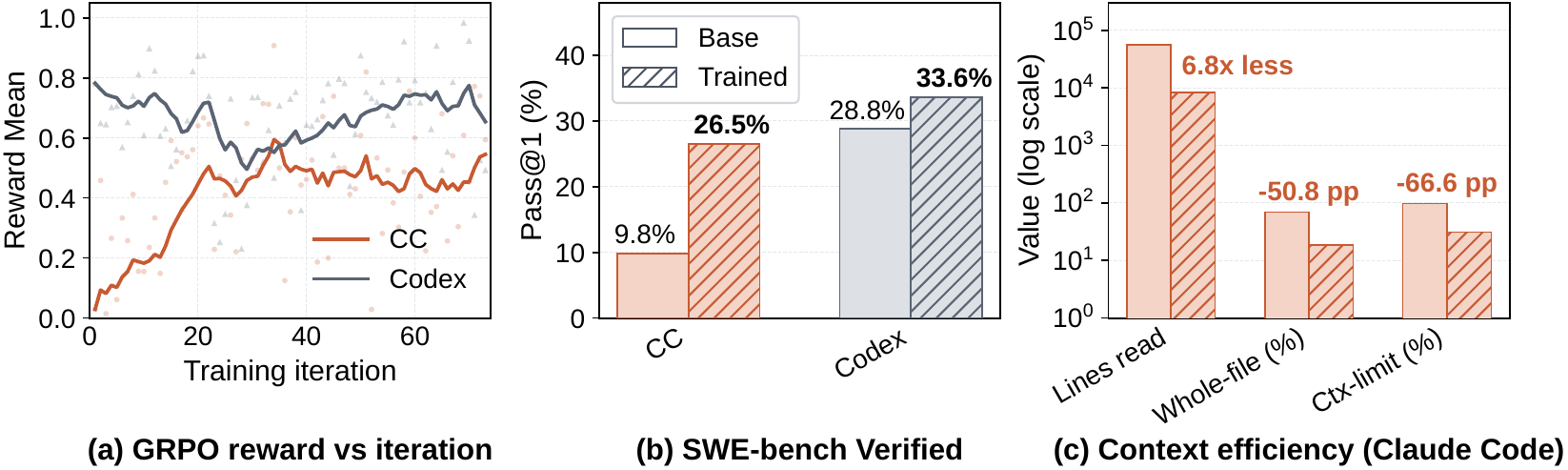}
\vspace{-15pt}
\caption{
SWE-Gym post-training and downstream SWE-bench Verified generalization on \aaw{}.
(a) GRPO rollout mean reward across 73 training iterations with exponential moving averages ($\alpha{=}0.15$).
(b) Pass@1 on SWE-bench Verified comparing base and post-trained checkpoints across harnesses.
(c) Context use under Claude Code on 42 matched tasks, where the trained model reads fewer lines, avoids whole-file reads, and rarely hits the context limit.
}
\label{fig:swegym-training}
\vspace{-8pt}
\end{figure*}

\noindent \textbf{Multi-session post-training on MetaClaw.} \label{sec:train-metaclaw}
We next test whether post-training through the unmodified harness and memory teaches the model to use them.
We post-train a Qwen3.5-4B model~\citep{qwen3.5} on the 30-workday MetaClaw benchmark by on-policy distillation (OPD) under Claude Code and Hermes, each with no memory, Mem0, and Hindsight.
The student generates every rollout through the unmodified harness and is trained on the per-token log-probability gap to a Qwen3.5-35B-A3B teacher that is given the round's corrective feedback, which the student never sees (Appendix~\ref{app:opd}).
Under Hindsight, the memory system's fact-extraction calls also run on the student and pass through the capture proxy, so they are trained alongside the agent's own turns and post-training shapes both how the model acts and what it writes to memory.
We train on simulated days 1--20 and evaluate base and trained checkpoints on all 346 rounds (Table~\ref{tab:sclate-training-metaclaw}).
On the held-out days 21--30, accuracy rises in all six configurations, by 7.6 to 11.8 pp in five of them, and the gain is significant ($p<0.05$) in three.
We report the change from base to the final epoch 5, and Figure~\ref{fig:opd-epoch} in Appendix~\ref{app:opd} shows held-out accuracy at every epoch.

We also train Hermes with Mem0 on the first two-thirds of each day's rounds and evaluate on the last third, to test whether post-training improves unseen rounds without the shift to new, harder days that the day split introduces.
On these 104 rounds, accuracy rises by 20.2 pp ($p = 0.007$, Appendix~\ref{app:tailcut}).

\noindent \textbf{Training changes how the model uses its memory system.}
\label{sec:qual-memwrite}
We compare the memory writes of the base and epoch-5 trained checkpoints on Hermes with Hindsight to see what the model does differently, beyond what accuracy shows (Appendix~\ref{app:memwrite}).
After every agent turn, Hindsight uses the same model as the agent to extract facts from the Hermes transcript into a fixed schema (statement, date, \texttt{Involving} entities, optional purpose clause), so the facts reflect the model being trained.
The trained model writes longer facts, and nearly all of the added length is metadata, which doubles from 36 to 76 characters while statements stay near 100 (Figure~\ref{fig:memwrite-isolation}a).
It also fills the purpose clause, stating why a fact was worth recording, in $19.6\%$ of facts versus $1.1\%$ before (Figure~\ref{fig:memwrite-isolation}b).
Facts naming any of the benchmark's five procedural rules rise from $23.6\%$ to $40.6\%$ (Figure~\ref{fig:memwrite-isolation}b), a gap that holds at every fact length (Figure~\ref{fig:memwrite-mechanism}a in Appendix~\ref{app:memwrite}).
Facts whose \texttt{Involving} field names the user fall from $65.3\%$ to $31.5\%$ while those naming the agent rise from $15.7\%$ to $47.0\%$, so the trained model records what the agent did where the base model records what it was told.
Appendix~\ref{app:memwrite} gives length-matched controls, per-convention breakdowns, and verbatim examples.

\noindent \textbf{Single-session post-training on SWE-Gym.}
We also validate \aaw{}'s post-training by executing GRPO~\citep{shao2024deepseekmath} on SWE-Gym paired with the same Qwen3.5-4B model across unmodified Claude Code and Codex harnesses, where the in-container capture proxy records the model's tokens and the grader's verdicts become GRPO rewards.
Over 73 iterations, the Claude Code rollout reward rises from 21.4\% to 53.2\% (first versus last ten iterations), while Codex stays near 69\% (Figure~\ref{fig:swegym-training}a).
The trained models transfer to SWE-bench Verified, lifting Pass@1 from 9.8\% to 26.5\% under Claude Code and from 28.8\% to 33.6\% under Codex (Figure~\ref{fig:swegym-training}b).
On 42 matched evaluation tasks, the trained Claude Code model reads $6.8\times$ fewer lines, cuts whole-file reads from 69.2\% to 18.4\%, and hits the context limit on 31.0\% of tasks instead of 97.6\% (Figure~\ref{fig:swegym-training}c).

\section{Conclusion}

We presented \aaw{}, an execution substrate built on an open event scheduler that lets benchmarks and agents place their events on one simulated timeline, so each needs only its own adapter instead of a custom scheduling loop for every pair.
The same scheduler runs month-long MetaClaw and asynchronous GAIA-2 scenarios without benchmark-specific scheduling logic, and its hybrid clock compresses 40 calendar days into a median of 5.9 hours with events kept on schedule and in order.
Comparing ten harness and memory configurations across ten models shows that no harness or memory is best for every model and that models differ widely in how they use the same harness and memory, so continual-learning agents must be evaluated with all three together.
Finally, \aaw{} enables post-training a model through its unmodified harness and memory, and the trained model adapts to both, working within its harness's context limit and writing richer memory records.

% \subsection*{Reproducibility statement}
% Complete environment specifications and instructions for reproducing the MetaClaw evaluation sweeps and the SWE-Gym post-training experiments are provided in Section~\ref{sec:framework}, Section~\ref{sec:experiments}, and Appendix~\ref{sec:reproducibility}.

\subsection*{AI use statement}
Generative AI tools were used to assist with coding and edit the paper for grammar, clarity, and concision.
All experimental implementations, systems designs, mathematical formulations, and textual claims were verified and validated by the authors, who take full responsibility for the contents of this paper.

\bibliographystyle{iclr2027_conference}
\bibliography{aaw_paper}

\newpage
\appendix

\section{Comparison with Related Systems} \label{app:related_comparison}

Table~\ref{tab:related-capabilities} compares \aaw{} with the evaluation and post-training systems discussed in Section~\ref{sec:related}, with every cell for another system taken from that system's own paper.

\begin{table}[t]
\caption{Requirements for evaluating and training a continual-learning agent as deployed, ordered so that each depends on those to its left.
Long horizons need multi-session scenarios and time acceleration, an accelerated clock must be shared by the benchmark and the agent, and agent-side events, such as session stop and start, state save and restore, crons, and memory consolidation, must then run on that shared clock, scheduled alongside the benchmark's events.
\cmark{} supported, \pmark{} partial, \xmark{} not supported or not described in the system's paper or documentation. 
A partial shared clock is real wall-clock time; GAIA-2 evaluates only its own scaffold, and rLLM agents are written against its own agent interface; MetaClaw runs only OpenClaw with a rewritten prompt, carries skills but not agent processes across workdays, and schedules its own training in idle windows on real time.}
\label{tab:related-capabilities}
\centering
\scriptsize
\setlength{\tabcolsep}{4pt}
\renewcommand{\arraystretch}{1.15}
\begin{tabular}{@{}l cc cc cc@{}}
\toprule
& \multicolumn{2}{c}{\textbf{As deployed}} & \multicolumn{2}{c}{\textbf{Long horizon}} & \multicolumn{2}{c}{\textbf{Faithful execution}} \\
\cmidrule(lr){2-3} \cmidrule(lr){4-5} \cmidrule(lr){6-7}
\textbf{System} & Harness & Memory & Multi-session & Time accel. & Shared clock & Agent-side events \\
\midrule
\multicolumn{7}{@{}l}{\textit{Evaluation}} \\
Harbor~\citep{harbor2026}              & \cmark & \xmark & \xmark & \xmark & \pmark & \xmark \\
GAIA-2 / ARE~\citep{froger2026gaia2}   & \pmark & \xmark & \xmark & \pmark & \pmark & \xmark \\
MetaClaw~\citep{xia2026metaclaw}       & \pmark & \xmark & \pmark & \xmark & \xmark & \pmark \\
\midrule
\multicolumn{7}{@{}l}{\textit{Post-training}} \\
POLAR~\citep{xu2026polar}              & \cmark & \xmark & \xmark & \xmark & \pmark & \xmark \\
OpenForgeRL~\citep{yu2026openforgerl}  & \cmark & \xmark & \xmark & \xmark & \pmark & \xmark \\
Agent Lightning~\citep{he2026agent}    & \cmark & \xmark & \xmark & \xmark & \pmark & \xmark \\
rLLM~\citep{tan2025rllm}               & \xmark & \xmark & \xmark & \xmark & \pmark & \xmark \\
\midrule
\textbf{\aaw{} (ours)}                 & \cmark & \cmark & \cmark & \cmark & \cmark & \cmark \\
\bottomrule
\end{tabular}
\end{table}

\section{Interface Contract Specifications} \label{sec:appendix_interfaces}

This section details the plug-in interfaces that decouple external workloads from \aaw{}'s execution core (Table~\ref{tab:interfaces}), formalizes the open event scheduling algorithm (Algorithm~\ref{alg:sclate_scheduling_loop}), details the shared event source catalog (Table~\ref{tab:event_sources}), and summarizes the typed effect primitives executed by the Effect Runner (Table~\ref{tab:effects}).

\begin{table*}[t]
\caption{Plug-in interface contracts in \aaw{}.
The substrate communicates with benchmark adapters, agent adapters, and event sources through uniform interfaces, decoupling environment definitions and agent harnesses from the open event scheduler.
An Agent Adapter covers a harness together with its memory system, so it also declares how their durable state is saved and restored. Every turn is delivered without blocking, so events can reach the agent while a turn is still running.}
\label{tab:interfaces}
\centering
\footnotesize
\renewcommand{\code}[1]{\texttt{#1}}% local: inherit the column's \scriptsize instead of \code's \small
\renewcommand{\arraystretch}{1.08}
\begin{tabularx}{\linewidth}{>{\scriptsize}p{4.5cm}X}
\toprule
{\footnotesize\textbf{Method \& Signature}} & \textbf{Systems Semantics} \\
\midrule
\multicolumn{2}{l}{\textit{\textbf{BenchmarkAdapter} $\longleftrightarrow$ Substrate (Scenario Environment and Evaluation)}} \\
\code{setup(scenario, state\_dir)} & Provisions initial mock application state (mail, calendar, git). \\
\code{source\_manifest(ctx)} / \code{event\_sources(ctx)} & Yields scenario tasks, inter-session world updates, session boundaries, and grading, either as declarations the scheduler instantiates or as ready-made sources. \\
\code{world\_provider()} & Optionally exposes a read-only view of application state as $\mathrm{ctx}$.world to event sources and graders. \\
\code{grading\_policy()} & Declares which turns are graded and when, how turns that are not graded are discharged, and whether a graded turn may be regraded, so the scheduler carries no benchmark-specific grading logic. \\
\code{Grader.evaluate(ctx) -> Judgment} & Independent oracle that scores each turn the grading policy selects against the task rubric, reading the dispatch context $\mathrm{ctx}$ including the world state, and returns the verdict and feedback. \\
\midrule
\multicolumn{2}{l}{\textit{\textbf{Agent Adapter} $\longleftrightarrow$ Substrate (Harness and Memory Lifecycle, Message Delivery, State Save and Restore)}} \\
\code{register\_capabilities(sources, prompts, hooks)} & Declares the agent's own event sources, such as memory consolidation and crons, together with its session-start and handoff prompts and hooks, which the scheduler merges onto the shared timeline. \\
\code{is\_ready() -> bool} & Polled readiness probe for agent process and memory sidecar initialization. \\
\code{end\_session(session\_id)} & Gracefully ends the session at a boundary, flushing session state to its durable store. \\
\code{shutdown()} & Whole-run teardown that stops the agent process and memory sidecars and releases their sockets. \\
\code{notify(turn)} & Delivers a user or environment prompt from the scheduler into the live agent session without blocking, opening a session if none is open. \\
\code{drain(settle) -> list[Reaction]} & Collects and parses the turns the agent completed, including reactions to notifications and native cron fires; each end of turn marks the agent idle for the clock. \\
\code{sniff\_wakeups() -> list[Wakeup]} & Reports the crons the agent registered in its own schedule store, whose fire times the scheduler computes. \\
\code{turn\_in\_flight() -> bool} & True while a pushed turn has not reached its end of turn, so the runner never grades a turn before the agent finishes acting. \\
\code{MemorySpec(agent\_dirs, memory\_dirs, hooks)} & Declares separately where the harness and its memory system keep durable state, with optional hooks that customize the snapshot and restore. \\
\code{save\_state(scope, dest)} & Saves the agent state, the memory-system state, or both, as selected by \code{scope}, as a directory snapshot or a sidecar export with an integrity token, backing the SaveState effect. \\
\code{restore\_state(scope, src)} & Restores the selected state from a directory snapshot or a sidecar export and refuses a torn or partial save, backing the RestoreState effect. \\
\midrule
\multicolumn{2}{l}{\textit{\textbf{ScheduleSource} $\longleftrightarrow$ Open Scheduler (Uniform Event Protocol)}} \\
\code{next\_fire(ctx) -> Fire | None} & Returns the next candidate fire and its effects, or none if the source is dormant. \\
\code{depends\_on: tuple[source\_id]} & Sources whose fires must complete before this source becomes eligible (causal gating). \\
\bottomrule
\end{tabularx}
\end{table*}

\noindent \textbf{Event source mechanics and scheduling algorithm.}
Algorithm~\ref{alg:sclate_scheduling_loop} formalizes the execution sequence of \aaw{}'s open event scheduler across its four stages.
In Stage~1, the scheduler synthesizes the immutable \code{DispatchContext} snapshot from current simulated time, observable world state, recent agent actions, evaluation judgments, and the completed fires registry ($\mathcal{H}$).
In Stage~2, the \code{FrontierQueue} polls each registered \code{ScheduleSource} via \code{next\_fire(ctx)}.
A source dynamically generates candidate fires ($f$) by evaluating temporal deadlines ($c \ge t$) or causal predicates over $\mathcal{H}$, returning $\bot$ if dormant.
In Stage~3, the time controller selects the earliest ready fire $e^\star$ based on scheduling precedence, allowing time to flow tick-by-tick while active and jumping across empty intervals when settled.
In Stage~4, the \code{EffectRunner} dispatches the orthogonal effect primitives declared in $e^\star.\text{effects}$ across filesystem IPC channels, appending $e^\star.\text{fire\_id}$ to $\mathcal{H}$ for downstream causal gating.

\begin{algorithm}[t]
\caption{\aaw{} Open Event Scheduling and State Synchronization Loop}
\label{alg:sclate_scheduling_loop}
\begin{algorithmic}[1]
\Require Registered event sources $\mathcal{S}$, simulated clock $c$, Effect Runner $\mathcal{E}$
\State $\mathcal{H} \gets \emptyset$ \Comment{Completed event fires registry (\code{completed\_fires})}
\While{$\text{alive}(\mathcal{S}) \lor \mathrm{ctx}.\text{is\_busy}$}
  \State $\mathrm{ctx} \gets \text{DispatchContext}(c, \mathcal{W}, \mathcal{A}, \mathcal{J}, \mathcal{H}, \text{activity})$ \Comment{Stage 1: Synthesize snapshot}
  \State $\mathcal{F} \gets \bigcup_{s \in \mathcal{S}} \{s.\mathrm{next\_fire}(\mathrm{ctx})\} \setminus \{\bot\}$ \Comment{Stage 2: Poll candidate fires from active sources}
  \If{$\mathrm{ctx}.\text{is\_busy}$} \Comment{Stage 3: Mode switch --- active agent reasoning or tool execution}
    \State $c \gets c + \Delta t_{\mathrm{wall}}$ \Comment{Flow mode: advance simulated clock with wall-clock time}
    \State $\mathcal{F}_{\mathrm{ready}} \gets \{f \in \mathcal{F} \mid f.t \le c\}$ \Comment{Filter candidate fires ready at or before current time}
    \If{$\mathcal{F}_{\mathrm{ready}} \neq \emptyset$}
      \State $e^\star \gets \arg\min_{f \in \mathcal{F}_{\mathrm{ready}}} (f.t, f.\text{source\_id})$ \Comment{Select earliest ready fire (tie-break on source\_id)}
      \State $\mathcal{E}.\text{execute}(e^\star.\text{effects})$ \Comment{Stage 4: Dispatch effect primitives}
      \State $\mathcal{H} \gets \mathcal{H} \,\|\, e^\star.\text{fire\_id}$
    \EndIf
  \Else \Comment{Jump mode: agent and sidecars are settled (idle)}
    \If{$\mathcal{F} = \emptyset$}
      \State \textbf{break} \Comment{All scenario and agent sources exhausted}
    \EndIf
    \State $e^\star \gets \arg\min_{f \in \mathcal{F}} (f.t, f.\text{source\_id})$ \Comment{Select earliest scheduled candidate}
    \State $c \gets \max(c, e^\star.t)$ \Comment{Instantaneous leap to scheduled event timestamp}
    \State $\mathcal{E}.\text{execute}(e^\star.\text{effects})$ \Comment{Stage 4: Dispatch effect primitives}
    \State $\mathcal{H} \gets \mathcal{H} \,\|\, e^\star.\text{fire\_id}$
  \EndIf
  \State $\mathcal{W}, \mathcal{A}, \mathcal{J} \gets \text{drain\_ipc\_updates}()$ \Comment{Update observable state from filesystem IPC channels}
\EndWhile
\end{algorithmic}
\end{algorithm}

\noindent \textbf{Shared event source catalog.}
Table~\ref{tab:event_sources} summarizes the core event source implementations provided by \aaw{}, partitioned into benchmark-contributed, agent-contributed, and shared causal dependency sources.

\begin{table*}[t]
\caption{Shared event source catalog in \aaw{}.}
\label{tab:event_sources}
\centering
\footnotesize
\renewcommand{\arraystretch}{1.08}
\begin{tabularx}{\linewidth}{p{4.8cm}p{2.6cm}X}
\toprule
\textbf{Event Source Class} & \textbf{Contributor} & \textbf{Systems Responsibility and Fire Dynamics} \\
\midrule
\code{UserInputSource} & Benchmark & Yields multi-turn task prompts and user instructions dynamically as prior turns are graded. \\
\code{SessionStartSource} & Shared & Coordinates multi-phase session opens, advancing time, invoking \code{StartAgent}, and folding session-init turns. \\
\code{SessionEndSource} & Shared & Manages session closes in two stages: delivering the handoff notification, then invoking \code{ShutdownAgent} once settled. \\
\code{AppStateConditionedSource} & Benchmark & Polls \code{ctx.world} during active flow mode, firing immediately when benchmark application predicates are satisfied. \\
\code{GradingSource} & Benchmark & Schedules evaluation fires across turns and session boundaries, invoking the independent oracle grader. \\
\midrule
\code{CronSource} / \code{WakeupSource} & Agent & Emits recurring wakeup targets on the timeline to pace native agent background crons and periodic maintenance. \\
\code{MemoryConsolidationSource} & Agent & Schedules background memory indexing and synthesis routines when the agent becomes idle or after session handoffs. \\
\code{SidecarReadySource} & Agent & Evaluates memory system readiness, withholding initial task delivery until sidecar sockets initialize. \\
\midrule
\code{EventChainSource} & Shared & Gates execution on prior event completions, firing once all prerequisite \code{fire\_id}s appear in \code{ctx.completed\_fires}. \\
\code{AgentActionGatedSource} & Shared & Gates execution on agent tool behavior, firing once the agent executes designated actions recorded in \code{ctx.agent\_actions}. \\
\bottomrule
\end{tabularx}
\end{table*}

\noindent \textbf{Effect primitives catalog.}
When the Frontier Queue selects an eligible event fire, the Effect Runner executes it through an orthogonal set of typed effect primitives.
Table~\ref{tab:effects} details these primitives, formalizing their operational semantics across IPC channels, process lifecycles, and storage boundaries.

\begin{table*}[t]
\caption{Effect primitives in \aaw{}'s Effect Runner.}
\label{tab:effects}
\centering
\footnotesize
\renewcommand{\arraystretch}{1.08}
\begin{tabularx}{\linewidth}{p{3.6cm}X}
\toprule
\textbf{Effect Primitive} & \textbf{Operational Semantics and State Mutations} \\
\midrule
\code{Deliver} & Delivers structured task instructions, user prompts, or system notifications directly to the agent over filesystem IPC channels. \\
\code{RunHook} & Executes a custom callback defined by an adapter to mutate environment state, seed database tables, or evaluate grading rubrics. \\
\code{AsyncReserve} & Switches the scheduler into real-time flow mode to wait for an asynchronous hook to finish, advancing time continuously without blocking the scheduler loop. \\
\code{AwaitWakeup} & Pauses timeline progression until the agent process settles into an idle state, becomes alive, or triggers an internal wakeup alarm. \\
\code{StartAgent} & Spawns the agent harness process along with any configured memory system inside the execution sandbox. \\
\code{ShutdownAgent} & Sends graceful termination signals to shut down the running agent harness and sidecar processes before closing the session. \\
\code{SaveState(scope)} & Invokes the adapter's \code{save\_state}, saving the agent workspace, the memory-system state, or both, as selected by \code{scope}, as a directory snapshot or a sidecar export. \\
\code{RestoreState(scope)} & Invokes the adapter's \code{restore\_state}, reloading the selected agent and memory state before opening the next session horizon. \\
\bottomrule
\end{tabularx}
\end{table*}

\noindent \textbf{Example benchmark adapter.}
Figure~\ref{lst:gaia2-adapter} excerpts the GAIA-2 adapter from the \aaw{} implementation, with logging, error handling, and custom-event plumbing elided.
Through \code{event\_sources}, the adapter turns the upstream scenario's user task and every environment event of the oracle DAG into ready-made sources, attaching to each environment event an app-state mutation and a notification, while \code{source\_manifest} only declares its grading configuration.
Each environment event with a dependency becomes a \code{\_CompositeDepSource} whose \code{next\_fire} waits until the agent has performed the gating app actions and every parent event has fired, then schedules the event at the scenario's delay after the moment its gates first held, so the scheduler, not the adapter, decides when to flow and when to jump.

\begin{figure}[t]
\centering
\begin{lstlisting}[language=Python, basicstyle=\ttfamily\scriptsize, frame=single, columns=fullflexible, keepspaces=true, showstringspaces=false, commentstyle=\color{aawgray}, keywordstyle=\color{aawblue}\bfseries, xleftmargin=2pt, xrightmargin=2pt]
class Gaia2Adapter:
    def setup(self, scenario_path, state_dir):
        self._scenario_path = Path(scenario_path)
        self._loader = _load_scenario(scenario_path)        # upstream scenario JSON
        ...
    def event_sources(self, ctx):                          # ready-made sources
        return self._native_event_sources(ctx) + self._custom_event_sources(ctx)

    def _native_event_sources(self, ctx):
        def effect_factory(ev):                            # payload of one ENV event
            action = [(ev.app, ev.function, ev.args, ev.event_id)]
            return [MutateEnv(apply=lambda acts=action:
                              self._native_execute_env(acts, state_dir)),
                    Deliver(text="[A notification just arrived ...]\n\n"
                                 + self._native_env_note(ev),
                            starts_turn=False, ungraded=True)]
        plan = build_native_plan(self._scenario_path, start_time=start_time,
                                 effect_factory=effect_factory,
                                 system_prompt=self._agents_md)
        return plan.sources                                # USER turn + one per ENV

    def source_manifest(self, ctx):                        # declarations
        return [SourceDeclaration(cls=None, label="gaia2-grading-config",
                    config={"grading": {"require_reply": True,
                                        "freeze_clock": False, ...}})]

    def world_provider(self):                              # becomes ctx.world
        return Gaia2WorldView(str(get_benchmark().gaia2_state_dir), adapter=self)

    def grading_policy(self):                              # whole-run oracle grade
        return AggregateAtClosePolicy()

class _CompositeDepSource:                                 # one ENV event with deps
    def next_fire(self, ctx):
        if self.source_id in ctx.completed_fires:
            return None
        have = ctx.world.agent_action_counts()
        if any(have[fn] < c for fn, c in self._need.items()):
            return None                                    # agent has not acted yet
        if not all(dep in set(ctx.completed_fires) for dep in self._chain_after):
            return None                                    # parent events pending
        if self._satisfied_at is None:
            self._satisfied_at = ctx.now                   # latch first gate match
        when = self._satisfied_at + timedelta(seconds=self._delay_s)
        return ScheduledFire(when=when, source_id=self.source_id,
                             required=self._required, effects=self._effects)
\end{lstlisting}
\vspace{-6pt}
\caption{Excerpt of the GAIA-2 benchmark adapter in \aaw{}, with logging, error handling, and custom-event plumbing elided. An environment event with no dependency uses a \code{\_TimedEnvSource} that fires at the scenario start plus its offset instead.}
\label{lst:gaia2-adapter}
\end{figure}

\section{Scheduler Measurements and Walkthroughs} \label{app:scheduler}

\subsection{Scheduler Measurements} \label{app:scheduler_metrics}

Table~\ref{tab:scheduler-metrics} measures the scheduler over every counted final-eval run of GAIA-2 and MetaClaw, computed from the raw scheduler traces together with the notification logs and harness session transcripts of each run.
For GAIA-2 the population is the run counted for each of the 12{,}000 cell and scenario pairs, and for MetaClaw it is the 100 runs of Table~\ref{tab:metaclaw-matrix}.

\begin{table}[t]
\caption{\aaw{} scheduler measurements over every counted final-eval run, from the raw scheduler traces. Benchmark-side events are user turns, grading, and GAIA-2 environment events; agent-side events are session starts and stops, including the handoff and init turns, and GBrain's nightly consolidation. GAIA-2 scenarios are single-session, so their runs dispatch no agent-side events. The scenario-specified delay is checked on the 6{,}999 GAIA-2 environment events whose parent the judge matched.}
\label{tab:scheduler-metrics}
\centering\small
\begin{tabular}{lr}
\toprule
\multicolumn{2}{l}{\textbf{MetaClaw}} \\
Runs & 100 \\
Simulated span, median & 40.0 days \\
Real span, median & 5.9\,h \\
Simulated over real time (all runs combined) & 142$\times$ \\
Benchmark-side events dispatched & 77{,}700 \\
Agent-side events dispatched & 12{,}400 \\
\midrule
\multicolumn{2}{l}{\textbf{GAIA-2}} \\
Runs & 12{,}000 \\
Benchmark-side events dispatched & 37{,}065 \\
Environment events fired & 13{,}050 of 13{,}050 \\
Fired at the scenario-specified delay & 98.7\% \\
Notifications reaching Claude Code mid-turn & 36.0\% \\
\bottomrule
\end{tabular}
\end{table}

\subsection{Simulated Time Inside Harness and Memory Processes} \label{app:time_grounding}

\begin{figure*}[t]
\centering
\includegraphics[width=0.98\linewidth]{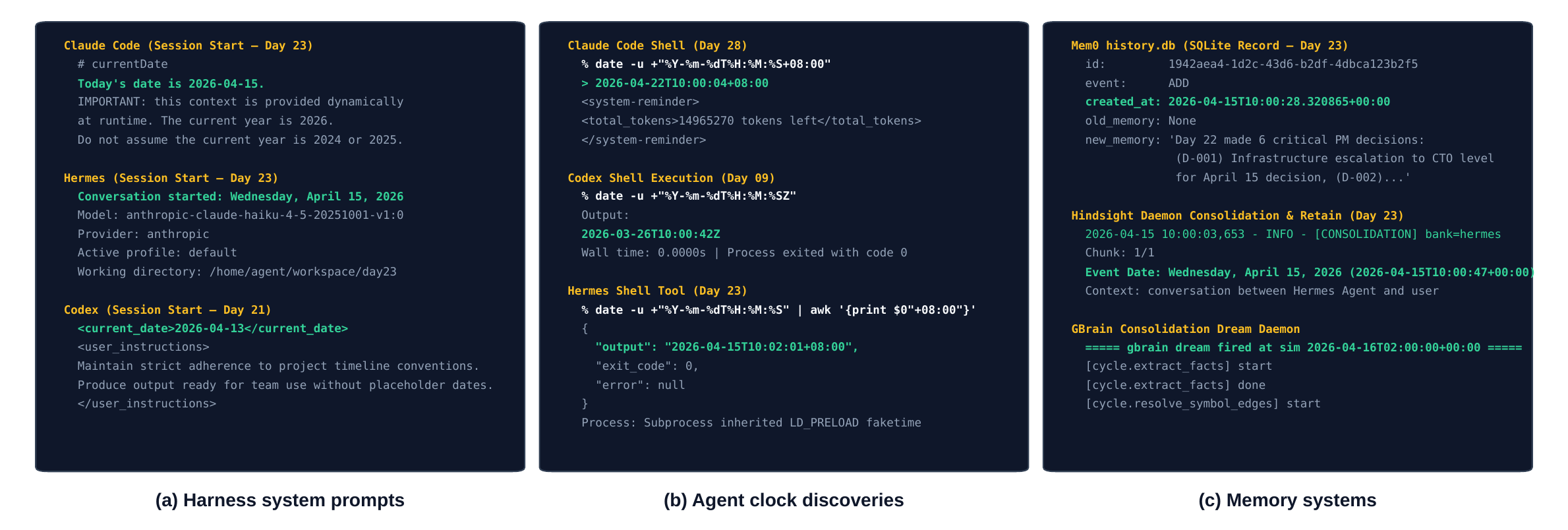}
\caption{Multi-layer simulated time grounding evidence across harnesses.
Verbatim excerpts from MetaClaw 30-workday evaluation runs showing synchronized simulated time across (a) harness system prompts injected at session boundaries, (b) active agent tool executions discovering time via shell commands under \code{libfaketime}, and (c) memory systems recording timestamped entries.}
\label{fig:time-grounding-evidence}
\end{figure*}

Figure~\ref{fig:time-grounding-evidence} shows the simulated clock reaching the harness and memory processes of MetaClaw runs.

\subsection{GAIA-2 Parity Grading} \label{app:gaia2_parity}

Upstream GAIA-2 judges each turn inside the run, at the agent's first message to the user, and ends the run at that judgment~\citep{froger2026gaia2}.
A rejected turn stops the scenario before its later events are delivered, and after the last event the run stops once the event queue drains, a median of 2.2\,s later and before the agent can answer it.
Its Hermes adapter also returns the text of an interrupted turn as the agent's message to the user, so a notification that arrives while Hermes waits on the model closes the turn with an intermediate response, as in about one in five upstream runs.
On a single-turn Time scenario these rules score only the actions taken before the agent's first progress message, so every upstream Time run fails even when the agent later completes the task on time.
For Figure~\ref{fig:gaia2-parity} we therefore remove the early stop from upstream GAIA-2, so a run ends only after the agent has answered the last environment event, and report an interrupted turn as interrupted rather than as a message to the user.
Every stack is then graded after the run ends on its whole trajectory by the same judge, the way \aaw{} grades GAIA-2 throughout the evaluation.

\subsection{Scheduler Walkthroughs on Real Trajectories} \label{app:scheduler_walkthroughs}

\begin{figure*}[t]
\centering
\includegraphics[width=\linewidth]{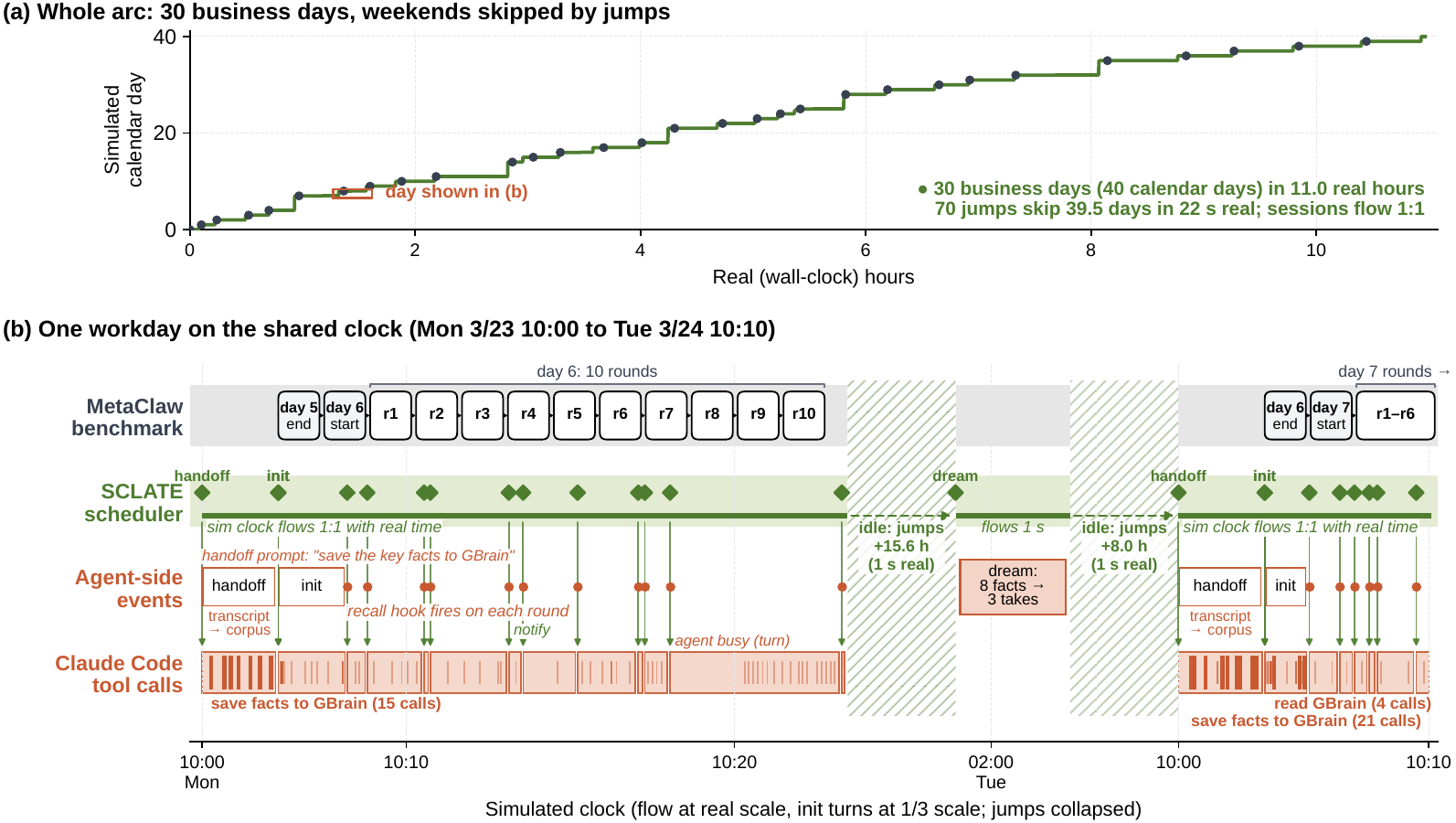}
\caption{MetaClaw under Claude Code with GBrain and Sonnet-5.
(a) Simulated calendar day versus real hours over the full arc, with one dot per work day.
(b) One work day on a broken time axis, where flow segments are drawn at real scale, init turns at one-third scale, and jumps collapsed into labeled breaks, showing benchmark rounds and agent-side events such as the handoff, the per-round recall hook, and GBrain's nightly consolidation on the same scheduler.}
\label{fig:walk-metaclaw}
\end{figure*}

\begin{figure*}[t]
\centering
\includegraphics[width=\linewidth]{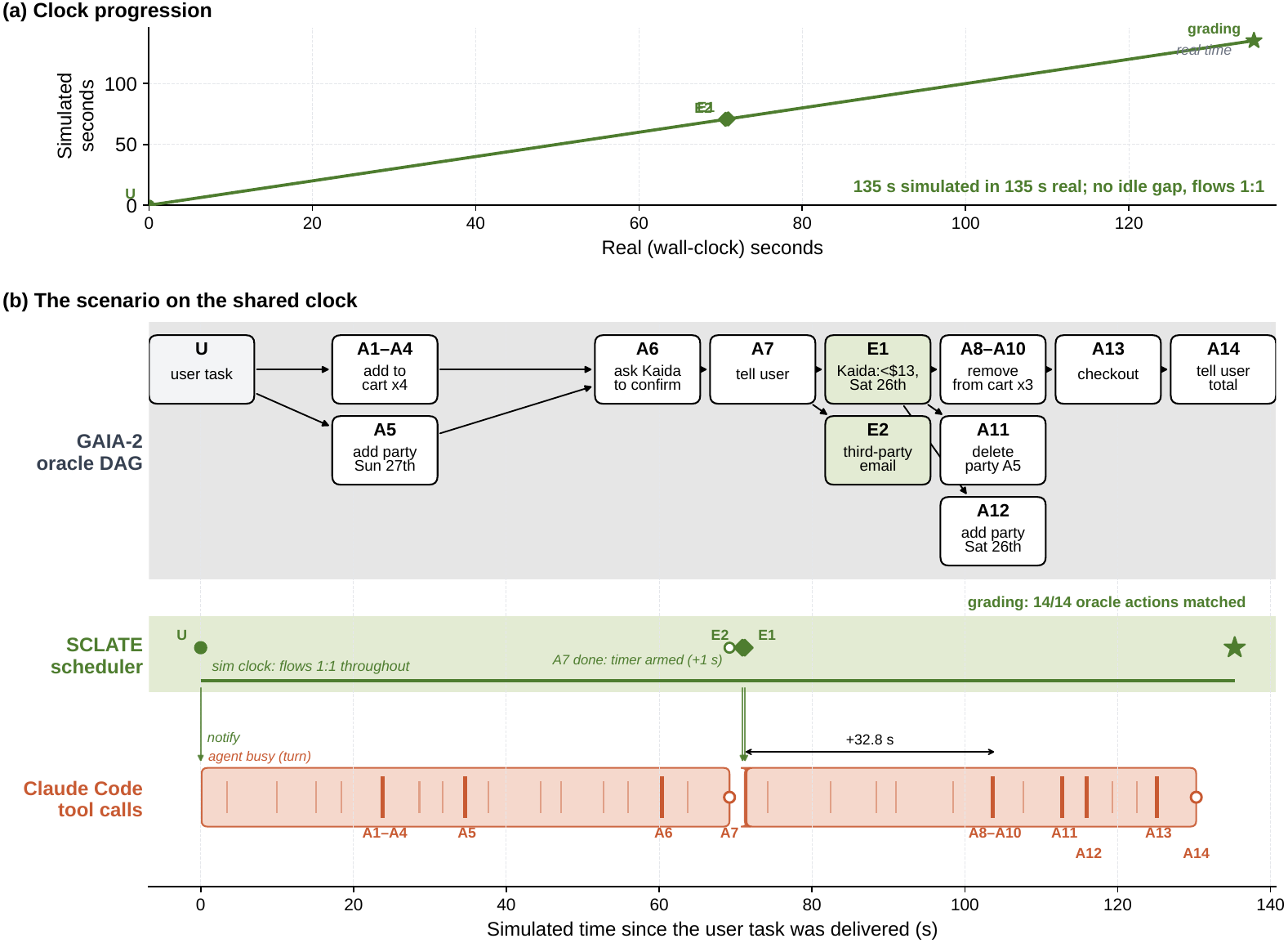}
\caption{GAIA-2 Adaptability scenario 155 under Claude Code with Sonnet-5.
(a) Simulated versus real time from the scheduler trace, flowing at real-time rate throughout.
(b) The oracle event DAG, the scheduler's dispatches and notifications, and the agent's turns and tool calls on one simulated time axis, where bold ticks are writes to application state labeled with the oracle node the judge matched.}
\label{fig:walk-gaia2-adapt}
\end{figure*}

\begin{figure*}[t]
\centering
\includegraphics[width=\linewidth]{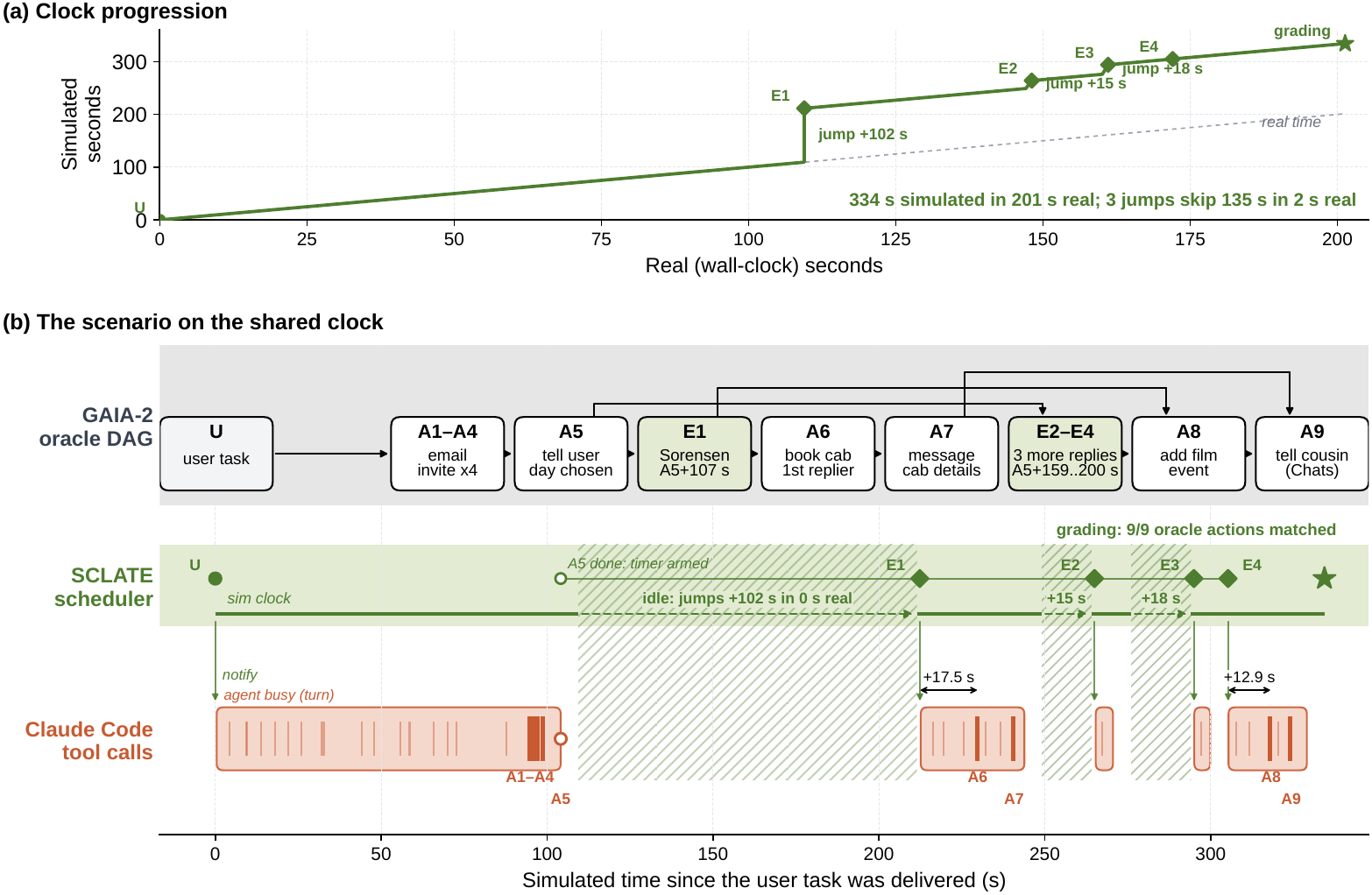}
\caption{GAIA-2 Time scenario 80 under Claude Code with Sonnet-5.
(a) Simulated versus real time, where the clock follows the real-time diagonal while the agent works and jumps vertically across idle gaps.
(b) The agent's message A5 arms timers for four scheduled replies, and the scheduler jumps the clock to each reply that falls due while the agent is idle, with hatched bands marking the jumps.}
\label{fig:walk-gaia2-time}
\end{figure*}

\begin{figure*}[t]
\centering
\includegraphics[width=\linewidth]{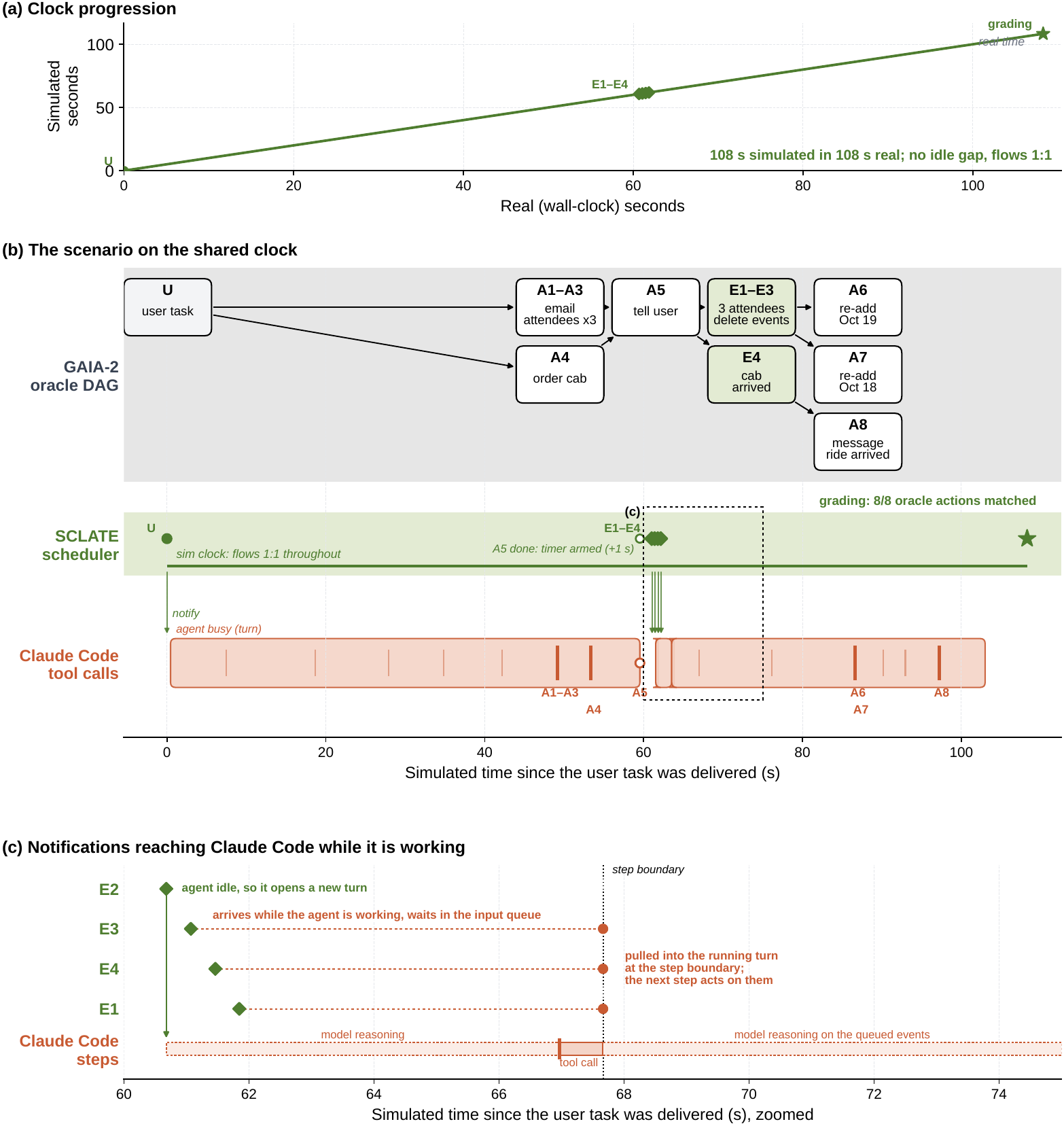}
\caption{GAIA-2 Adaptability scenario 148 under Claude Code with Sonnet-5.
(a) Simulated versus real time, flowing at real-time rate throughout.
(b) The oracle event DAG, the scheduler's dispatches, and the agent's turns and tool calls, with the dashed box marking the window shown in (c).
(c) Four events dispatched within 1.2 seconds, where the first opens a turn and the other three wait in Claude Code's input queue until the harness consumes them at its next step boundary.}
\label{fig:walk-gaia2-midturn}
\end{figure*}

Figures~\ref{fig:walk-metaclaw}--\ref{fig:walk-gaia2-midturn} trace how the \aaw{} scheduler executes benchmark scenarios on real trajectories, drawn directly from the raw artifacts of verified Claude Code runs with Sonnet-5, which include the scenario definition, the scheduler trace, the application-state log, the notification log, the agent session transcript, and the judgment.
Each figure aligns the benchmark, the scheduler, and the agent on one simulated time axis, and the MetaClaw figure adds a lane for agent-side events.
The top lane shows the benchmark scenario. For MetaClaw this is the chain of session boundaries and rounds, and for GAIA-2 it is the oracle event DAG, where \textbf{U} is the user task, \textbf{A} nodes are the actions the agent is expected to perform, and \textbf{E} nodes are environment events that the scenario itself applies to the world, such as an incoming reply email.
Every \textbf{A} node other than a message to the user is a write that mutates application state, and the judge matches these writes against the agent's own tool calls, while \textbf{E} nodes are world-state mutations that the scheduler, rather than the agent, performs.
The middle lane shows what the scheduler does. When the parent of an environment event is realized, the scheduler arms that event's timer at the delay the scenario specifies, dispatches the event when it falls due, and delivers a notification into the agent's session, while the simulated clock flows at real-time rate whenever the agent or a tool is active and jumps across settled idle intervals.
The bottom lane shows the agent, with each turn as a shaded box and each tool call as a tick, where bold ticks mark GBrain memory calls in MetaClaw and, in GAIA-2, the calls that wrote application state, labeled with the identifier of the oracle node the judge matched them to.
Panel (a) of each figure plots simulated time against real time from the same scheduler trace, so flow appears as the diagonal and each jump as a vertical step.

\noindent \textbf{Benchmark and agent-side events on one clock across days (MetaClaw).}
Figure~\ref{fig:walk-metaclaw} shows the mechanism over the 30-workday MetaClaw arc under Claude Code with GBrain.
The whole arc spans 40 calendar days in 11.0 real hours, as 70 jumps skip 39.5 simulated days, including every night and weekend, in 22 real seconds, while every session flows at real-time rate.
Sessions are contiguous, so each work day's session closes when the next one opens at 10:00, and the zoomed day begins with the handoff turn in which the harness prompts the agent to save the key facts of the previous day to GBrain, followed by an init turn that opens the new session.
The ten rounds of day 6 then flow in sequence, each dispatched by the scheduler, delivered as a notification that GBrain's prompt hook enriches with recalled memory, and graded before the next round is released.
When the last round settles, the scheduler jumps 15.6 hours in one real second to GBrain's nightly consolidation at 02:00, an agent-side event registered on the same scheduler as the benchmark's sessions, which promotes eight saved facts into three consolidated takes while the clock flows for its one-second run.
A further eight-hour jump reaches the next morning, where the handoff saves day 6's facts and day 7's init turn reads GBrain before its first round.

\noindent \textbf{Reacting to an event triggered by the agent's own action (GAIA-2 Adaptability).}
In Figure~\ref{fig:walk-gaia2-adapt}, the agent adds four product variants to the cart, creates a party event, asks the friend Kaida to confirm, and reports back to the user (A1--A7).
The report (A7) is the parent of two environment events, so the scheduler arms both at the scenario's one-second delay, delivering Kaida's reply that only items under \$13 and a Saturday date are wanted (E1) together with a third-party email that repeats the request as a distractor (E2).
The notification opens a new turn, and 32.8 simulated seconds later the agent removes the three disallowed items, moves the party to Saturday, checks out, and reports the total (A8--A14), acting on Kaida's message and ignoring the distractor.
In this scenario, the agent is never idle long enough for a jump, and the clock flows at real-time rate throughout.
Grading matches all 14 oracle actions, and only 3 of the 15 harness and model cells pass this scenario.

\noindent \textbf{Timed events across idle gaps (GAIA-2 Time).}
In Figure~\ref{fig:walk-gaia2-time}, the agent emails a film-night invite to four people (A1--A4) and tells the user which day it chose (A5), after which the scenario schedules the four replies at 107, 159, 189, and 200 seconds after that message (E1--E4).
Once the agent finishes its turn and nothing is in flight, the scheduler jumps the clock 102 simulated seconds in zero real seconds to the first reply, then alternates between flowing while the agent works and two short jumps of 15 and 18 seconds as the remaining replies fall due.
The agent books a cab for the first replier 17.5 seconds after the first reply (A6, A7) and creates the calendar event 12.9 seconds after the last reply (A8), both within the judge's time windows, and the judge matches all nine oracle actions.
The scenario spans 334 simulated seconds in 201 real seconds.

\noindent \textbf{Notifications arriving while the agent is working (GAIA-2 Adaptability).}
In Figure~\ref{fig:walk-gaia2-midturn}, the agent emails three event attendees asking whether they want to delete their coffee break, orders a cab, and reports to the user (A1--A5), and that report arms four environment events at the scenario's one-second delay, in which three attendees delete their events (E1--E3) and the cab arrives (E4).
The first of them, E2, reaches the agent while it is idle and opens a new turn, and the other three are dispatched within the next 1.2 seconds while the agent is already reasoning on it.
Claude Code places these three notifications in its own input queue and, when its first tool call returns about seven seconds later, pulls all three into the running turn, so its next step acts on the remaining deletions and the cab arrival together (Figure~\ref{fig:walk-gaia2-midturn}c).
Across all counted runs, 36.0\% of GAIA-2 notifications reach Claude Code inside a running tool loop in this way (Table~\ref{tab:scheduler-metrics}), and the judge matches all eight oracle actions of this scenario.

\section{Benchmarks and Evaluation} \label{sec:reproducibility}

\subsection{Benchmark Workload Specifications and Evaluation Protocol} \label{sec:benchmark_specs}

Table~\ref{tab:benchmark-specs} provides complete dataset specifications, evaluation splits, and task horizons for all seven ported benchmarks in \aaw{}, and Table~\ref{tab:reasoning-effort} lists the reasoning effort each model runs with.

\begin{table}[t]
\caption{Reasoning effort per model, the same under every harness.}
\label{tab:reasoning-effort}
\centering
\footnotesize
\setlength{\tabcolsep}{6pt}
\renewcommand{\arraystretch}{1.12}
\begin{tabular}{ll}
\toprule
\textbf{Model} & \textbf{Reasoning effort} \\
\midrule
Haiku-4.5 & high \\
Sonnet-5 & high \\
Opus-4.6 & high \\
Opus-4.8 & high \\
Gemini-3.7F & provider default \\
GPT-5.4 & none \\
GLM-5.3F & high \\
Qwen3.8-FN & medium \\
DS-Vision & medium \\
DS-V4.1 & high \\
\bottomrule
\end{tabular}
\end{table}

\begin{table*}[t]
\caption{Comprehensive specifications of the seven ported benchmark workloads evaluated in \aaw{}. Subsampled recall suites evaluate the first $N$ items of the listed split, and every arm of a benchmark evaluates the same items.}
\label{tab:benchmark-specs}
\centering
\footnotesize
\setlength{\tabcolsep}{3pt}
\renewcommand{\arraystretch}{1.12}
\begin{tabular*}{\linewidth}{@{\extracolsep{\fill}}llll@{}}
\toprule
\textbf{Benchmark} & \textbf{Upstream Size} & \textbf{Evaluated Tasks ($N$)} & \textbf{Task Domain \& Environment} \\
\midrule
\shortstack[l]{MetaClaw \\ \citep{xia2026metaclaw}} & \shortstack[l]{346 tasks \\ (30 workdays)} & All 346 tasks & Multi-session developer workspace \\
\addlinespace[3pt]
\shortstack[l]{GAIA-2 \\ \citep{froger2026gaia2}} & \shortstack[l]{800 tasks \\ (5 categories)} & All 800 tasks & Asynchronous personal assistant apps \\
\addlinespace[3pt]
\shortstack[l]{AppWorld \\ \citep{trivedi2024appworld}} & \shortstack[l]{750 tasks \\ (585 test)} & \shortstack[l]{All 585 test tasks \\ (168 normal, 417 challenge)} & Multi-app transactions \\
\addlinespace[3pt]
\shortstack[l]{LongMemEval \\ \citep{wu2024longmemeval}} & \shortstack[l]{500 tasks \\ (oracle/s/m)} & \shortstack[l]{100 tasks \\ ($s$ split)} & Multi-session conversational recall \\
\addlinespace[3pt]
\shortstack[l]{PersonaMem \\ \citep{jiang2025know}} & \shortstack[l]{5{,}990 tasks \\ (32k--1M)} & \shortstack[l]{100 tasks \\ (128k tier)} & Long-horizon persona tracking \\
\addlinespace[3pt]
\shortstack[l]{SWE-Gym \\ \citep{pan2024swegym}}  & 2{,}438 tasks & \shortstack[l]{293 tasks \\ (SkyRL-v0 subset)} & GitHub pull request resolution \\
\addlinespace[3pt]
\shortstack[l]{SWE-bench Verified \\ \citep{jimenez2024swebenchlanguagemodelsresolve}} & 500 tasks & All 500 tasks & Downstream code patch generation \\
\bottomrule
\end{tabular*}
\end{table*}

\noindent \textbf{Benchmark porting methodology.}
We port all seven benchmarks to \aaw{} under uniform execution semantics while strictly preserving upstream task definitions and ground truth.
For MetaClaw, upstream execution lacked an event scheduler and simulated temporal progression solely through prompt-injected date strings.
We ported MetaClaw to a real simulated clock visible across system calls, executing the agent and memory directly on this physical timeline so that real timestamps, file modification times, and cache TTLs are genuinely visible to agent processes.
For GAIA-2 and AppWorld, application webservers deploy directly inside each rollout container.
These servers are exposed to the agent exclusively through CLI tools configured with the \code{setuid} bit.
This \code{setuid} mechanism enforces Linux privilege separation, so the unprivileged \code{agent} user cannot access application webservers directly and must execute through the CLI tool, isolating the agent from the \code{daemonsched} daemon user that manages grading, evaluation, and event scheduling.
Each benchmark defines a \code{WorldProvider} adapter, allowing grading to inspect ground-truth environment state without leaking it to the agent.
For recall benchmarks (LongMemEval, PersonaMem), upstream suites lacked multi-session retain, consolidate, and recall lifecycles, evaluating static retrieval on pre-collected traces.
We grouped conversational sessions into framework sessions, scheduling each framework session on a distinct simulated day.
This temporal progression places an inter-session interval between days, across which the agent process restarts and earlier facts survive only through native memory or a memory system, while the underlying dataset content, questions, and ground-truth answers remain strictly preserved.

\noindent \textbf{Dataset subsampling and split design.}
For benchmarks where full upstream evaluation would require excessive compute, we evaluate a fixed subset.
For MetaClaw, GAIA-2, and AppWorld, we evaluate the full public benchmark suites without subsampling.
Specifically, MetaClaw evaluates the complete 30-workday arc spanning 346 rounds across progressive development sprints, GAIA-2 evaluates all 800 scenarios, 160 in each of its five capability categories (Execution, Search, Adaptability, Ambiguity, Time), under the unified simulated clock, and AppWorld evaluates all 585 tasks across both the normal (indices 0--167) and challenge (indices 168--584) test sets.
For long-horizon memory recall suites, we evaluate the first 100 items of one upstream split.
LongMemEval is evaluated on the first 100 items of the $s$ split, whose multi-session conversation history spans about 115k tokens per instance.
PersonaMem is evaluated on the first 100 items of the 128k-token tier, with dialog structured directly within user messages.
Every arm of a given benchmark evaluates the identical matched index set, so that differences between arms do not come from different items.
For post-training, reinforcement learning models train on the 293-task SkyRL-v0 subset of SWE-Gym (the train split of \code{NovaSky-AI/SkyRL-v0-293-data}), with downstream generalization evaluated on the full 500 instances of SWE-bench Verified.
The $N=42$ instance split reported in Section~\ref{sec:training_results} represents a matched evaluation subset used to measure context-window efficiency and targeted reading behaviors under identical task seeds.

\noindent \textbf{Faithful execution and scoring protocol.}
To keep comparisons across agent configurations rigorous, \aaw{} enforces a strict faithful scoring protocol.
First, to prevent artificial score inflation from multiple retries, canonical accuracy reflects the outcome of the latest faithful run per scenario rather than an optimistic best-of-$k$ metric.
Second, reasoning effort is matched per model across all driving harnesses (Table~\ref{tab:reasoning-effort}), ensuring that cross-backend comparisons remain unconfounded by varying model inference parameters.

\subsection{Evaluation Details and Tool Usage Analysis} \label{app:evaluation_details}

\begin{figure}[t]
\centering
\includegraphics[width=0.85\linewidth]{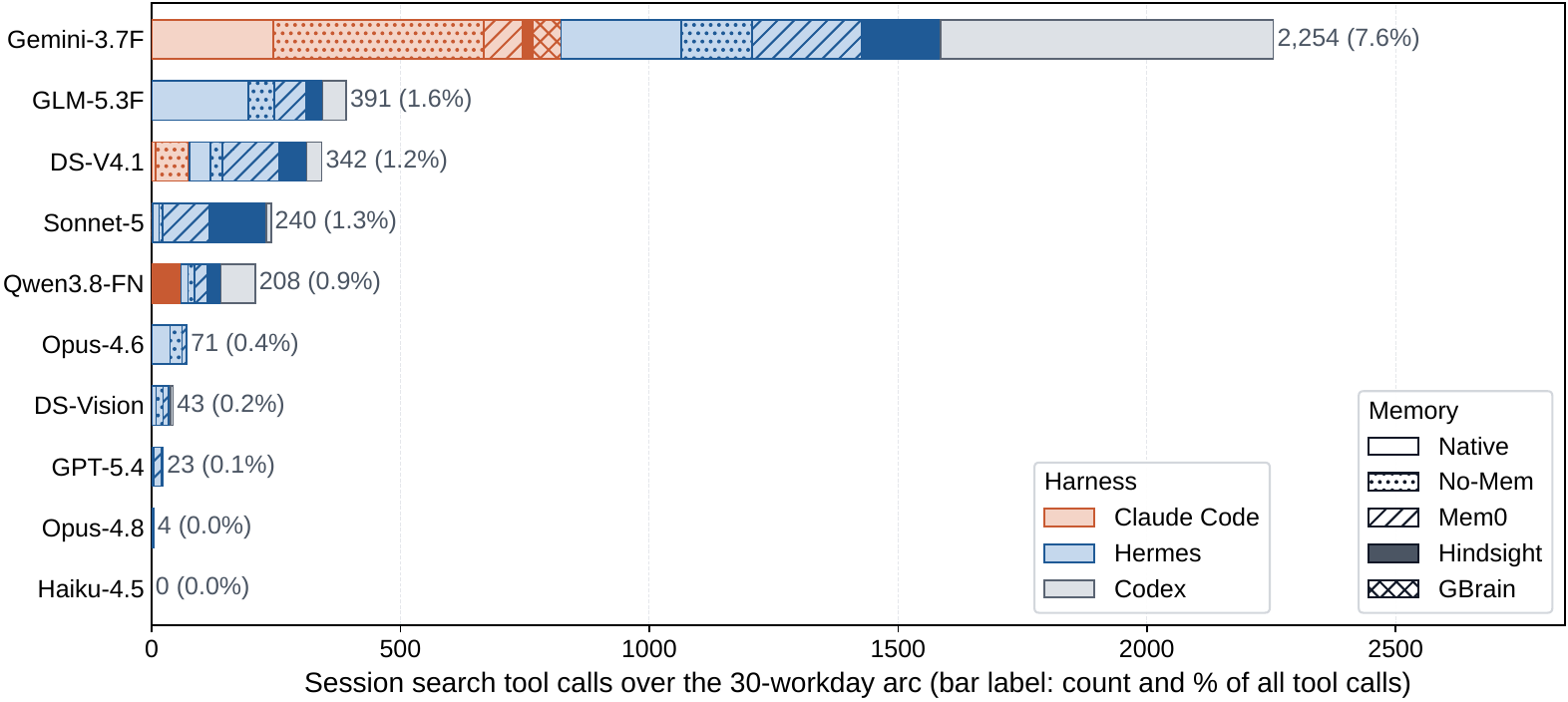}
\caption{Frequency of harness session-search tool invocations across models and driving harnesses on MetaClaw. Gemini-3.7F calls session search substantially more than other models across all harnesses, consistent with its elevated no-memory baseline performance.}
\label{fig:metaclaw-session-search}
\end{figure}

Figure~\ref{fig:metaclaw-session-search} quantifies the invocation frequency of harness session-search tools across models and driving harnesses on MetaClaw.
Across all configurations, Gemini-3.7F calls session search substantially more frequently than other models, actively inspecting prior workspace transcripts and recovering procedural rules even under the unaugmented baseline.

\noindent \textbf{Token usage across models and harnesses.}
Table~\ref{tab:metaclaw-tokens} reports total token consumption for every MetaClaw configuration of Table~\ref{tab:metaclaw-matrix}.
In this experiment, every model executes the identical 346-round timeline and these differences reflect model behavior rather than workload. 
We compare models by medians over the five configurations measured for all of them (Hermes No-Mem, Native, Mem0, and Hindsight, and Codex Native).
Gemini-3.7F consumes the most tokens (median 528M versus 95--242M for the other models, and the most in every configuration except Codex Native), driven by more agent calls per round (11.0) and a roughly twofold larger context per call (130K versus 38--69K tokens), consistent with its behavior of loading prior session transcripts.
GPT-5.4 is the most frugal, with the smallest contexts (38K) and shortest outputs (253 tokens per call).
Token consumption does not track accuracy, as DS-V4.1 comes within 4.1 pp of Gemini-3.7F's median accuracy with 0.32$\times$ its tokens (171M versus 528M), and Opus-4.6 uses fewer tokens than Haiku-4.5 (148M versus 242M) while scoring higher (65.0\% versus 59.8\%).
The harness shapes consumption as strongly as the model, as for every API-served model Hermes Native consumes 2.6--6.4$\times$ the tokens of Claude Code Native, because Hermes issues 2--4$\times$ more agent calls per round.

\begin{table*}[t]
\caption{
  MetaClaw total token usage (millions) over the 30-workday, 346-round arc for each model and harness--memory
  configuration of Table~\ref{tab:metaclaw-matrix}. 
  Totals count uncached input, cache reads, cache writes, and output across all agent and memory-system calls recorded by the \aaw{} proxy, which attributes each call to the agent or to the memory system itself. Across the 38 memory-backend cells with complete usage records, the memory system's own calls account for a median of 4.2\% of tokens, at most 13.8\% for Claude Code with Hindsight and under 0.3\% for GBrain. Cell color intensity scales with
  $\log$ tokens within each harness family. ``--'' marks Claude Code on self-hosted models. The proxy records these calls, but the self-hosted endpoint's streamed
  responses omit the usage fields this table sums.
}
\label{tab:metaclaw-tokens}
\centering
\scriptsize
\setlength{\tabcolsep}{2.5pt}
\renewcommand{\arraystretch}{1.20}
\begin{tabular*}{\linewidth}{@{\extracolsep{\fill}}lcccccccccc@{}}
\toprule
& \multicolumn{5}{c}{\footnotesize\textbf{Claude Code}} & \multicolumn{4}{c}{\footnotesize\textbf{Hermes}} & \multicolumn{1}{c}{\footnotesize\textbf{Codex}} \\
\cmidrule(lr){2-6} \cmidrule(lr){7-10} \cmidrule(lr){11-11}
\textbf{Model} & No-Mem & Native & Mem0 & Hindsight & GBrain & No-Mem & Native & Mem0 & Hindsight & Native \\
\midrule
Haiku-4.5 & \cellcolor{ccorange!8} 34 & \cellcolor{ccorange!9} 38 & \cellcolor{ccorange!18} 77 & \cellcolor{ccorange!16} 67 & \cellcolor{ccorange!21} 99 & \cellcolor{aawblue!13} 78 & \cellcolor{aawblue!26} 242 & \cellcolor{aawblue!27} 270 & \cellcolor{aawblue!26} 256 & \cellcolor{codexgray!14} 42 \\
Sonnet-5 & \cellcolor{ccorange!13} 51 & \cellcolor{ccorange!16} 67 & \cellcolor{ccorange!26} 159 & \cellcolor{ccorange!21} 104 & \cellcolor{ccorange!28} 179 & \cellcolor{aawblue!13} 76 & \cellcolor{aawblue!23} 197 & \cellcolor{aawblue!32} 442 & \cellcolor{aawblue!28} 310 & \cellcolor{codexgray!19} 70 \\
Opus-4.6 & \cellcolor{ccorange!11} 45 & \cellcolor{ccorange!14} 57 & \cellcolor{ccorange!23} 122 & \cellcolor{ccorange!21} 100 & \cellcolor{ccorange!22} 107 & \cellcolor{aawblue!10} 59 & \cellcolor{aawblue!20} 148 & \cellcolor{aawblue!26} 252 & \cellcolor{aawblue!24} 203 & \cellcolor{codexgray!12} 31 \\
Opus-4.8 & \cellcolor{ccorange!10} 41 & \cellcolor{ccorange!10} 42 & \cellcolor{ccorange!28} 176 & \cellcolor{ccorange!22} 109 & \cellcolor{ccorange!25} 144 & \cellcolor{aawblue!11} 65 & \cellcolor{aawblue!21} 166 & \cellcolor{aawblue!24} 209 & \cellcolor{aawblue!23} 198 & \cellcolor{codexgray!13} 34 \\
Gemini-3.7F & \cellcolor{ccorange!31} 227 & \cellcolor{ccorange!26} 155 & \cellcolor{ccorange!36} 349 & \cellcolor{ccorange!31} 228 & \cellcolor{ccorange!34} 295 & \cellcolor{aawblue!29} 317 & \cellcolor{aawblue!35} 588 & \cellcolor{aawblue!36} 625 & \cellcolor{aawblue!34} 528 & \cellcolor{codexgray!32} 322 \\
GPT-5.4 & \cellcolor{ccorange!6} 30 & \cellcolor{ccorange!7} 33 & \cellcolor{ccorange!16} 67 & \cellcolor{ccorange!15} 63 & \cellcolor{ccorange!19} 89 & \cellcolor{aawblue!6} 41 & \cellcolor{aawblue!15} 95 & \cellcolor{aawblue!19} 130 & \cellcolor{aawblue!18} 118 & \cellcolor{codexgray!6} 16 \\
GLM-5.3F & -- & -- & -- & -- & -- & \cellcolor{aawblue!12} 74 & \cellcolor{aawblue!24} 213 & \cellcolor{aawblue!32} 436 & \cellcolor{aawblue!26} 250 & \cellcolor{codexgray!23} 108 \\
Qwen3.8-FN & -- & -- & -- & -- & -- & \cellcolor{aawblue!9} 56 & \cellcolor{aawblue!24} 202 & \cellcolor{aawblue!25} 221 & \cellcolor{aawblue!22} 179 & \cellcolor{codexgray!36} 503 \\
DS-Vision & -- & -- & -- & -- & -- & \cellcolor{aawblue!10} 62 & \cellcolor{aawblue!20} 148 & \cellcolor{aawblue!23} 189 & \cellcolor{aawblue!20} 149 & \cellcolor{codexgray!16} 52 \\
DS-V4.1 & -- & -- & -- & -- & -- & \cellcolor{aawblue!14} 83 & \cellcolor{aawblue!22} 171 & \cellcolor{aawblue!25} 236 & \cellcolor{aawblue!26} 251 & \cellcolor{codexgray!17} 57 \\
\bottomrule
\end{tabular*}
\end{table*}

\subsection{Complete GAIA-2 Capability Breakdown} \label{app:gaia2_details}

Table~\ref{tab:gaia2-full-breakdown} provides the complete per-category Pass@1 (\%) results for all fifteen combinations of five models and three production harnesses across the five core GAIA-2 capability categories.

\begin{table*}[t]
\caption{
  Complete GAIA-2 Pass@1 (\%) breakdown across production harnesses and models on the full 800-scenario benchmark (160 scenarios per capability category).
  Cell color intensity scales continuously with accuracy within each harness family (terracotta for Claude Code, blue for Hermes, and slate for Codex).
  Bold indicates the highest accuracy for each model across harnesses.
}
\label{tab:gaia2-full-breakdown}
\centering
\scriptsize
\setlength{\tabcolsep}{8pt}
\renewcommand{\arraystretch}{1.20}
\begin{tabular*}{\linewidth}{@{\extracolsep{\fill}}llcccccc@{}}
\toprule
\textbf{Harness} & \textbf{Model} & \textbf{Execution} & \textbf{Search} & \textbf{Adaptability} & \textbf{Ambiguity} & \textbf{Time} & \textbf{Mean} \\
\midrule
Claude Code & Haiku-4.5 & \cellcolor{ccorange!8} 36.9\% & \cellcolor{ccorange!28} \textbf{70.0\%} & \cellcolor{ccorange!5} 18.8\% & \cellcolor{ccorange!4} 13.1\% & \cellcolor{ccorange!3} 9.4\% & \cellcolor{ccorange!9} 29.6\% \\
Hermes      & Haiku-4.5 & \cellcolor{aawblue!17} \textbf{53.8\%} & \cellcolor{aawblue!19} 56.2\% & \cellcolor{aawblue!5} 15.0\% & \cellcolor{aawblue!5} \textbf{14.4\%} & \cellcolor{aawblue!5} \textbf{13.8\%} & \cellcolor{aawblue!10} \textbf{30.6\%} \\
Codex       & Haiku-4.5 & \cellcolor{codexgray!10} 41.2\% & \cellcolor{codexgray!22} 62.5\% & \cellcolor{codexgray!5} 16.2\% & \cellcolor{codexgray!4} 12.5\% & \cellcolor{codexgray!3} 10.0\% & \cellcolor{codexgray!8} 28.5\% \\
\midrule
Claude Code & Sonnet-5   & \cellcolor{ccorange!31} \textbf{81.9\%} & \cellcolor{ccorange!36} \textbf{92.5\%} & \cellcolor{ccorange!14} \textbf{37.5\%} & \cellcolor{ccorange!28} \textbf{73.8\%} & \cellcolor{ccorange!11} 28.1\% & \cellcolor{ccorange!24} \textbf{62.8\%} \\
Hermes      & Sonnet-5   & \cellcolor{aawblue!28} 73.8\% & \cellcolor{aawblue!36} \textbf{92.5\%} & \cellcolor{aawblue!10} 26.9\% & \cellcolor{aawblue!23} 58.8\% & \cellcolor{aawblue!11} \textbf{29.4\%} & \cellcolor{aawblue!22} 56.2\% \\
Codex       & Sonnet-5   & \cellcolor{codexgray!29} 78.1\% & \cellcolor{codexgray!34} 86.9\% & \cellcolor{codexgray!11} 28.8\% & \cellcolor{codexgray!22} 56.9\% & \cellcolor{codexgray!5} 13.8\% & \cellcolor{codexgray!20} 52.9\% \\
\midrule
Claude Code & GLM-5.3F   & \cellcolor{ccorange!24} 63.8\% & \cellcolor{ccorange!30} 78.1\% & \cellcolor{ccorange!11} 28.8\% & \cellcolor{ccorange!9} 23.8\% & \cellcolor{ccorange!10} 27.5\% & \cellcolor{ccorange!17} 44.4\% \\
Hermes      & GLM-5.3F   & \cellcolor{aawblue!30} \textbf{78.1\%} & \cellcolor{aawblue!35} \textbf{90.0\%} & \cellcolor{aawblue!11} \textbf{30.0\%} & \cellcolor{aawblue!12} \textbf{31.2\%} & \cellcolor{aawblue!11} \textbf{28.8\%} & \cellcolor{aawblue!20} \textbf{51.6\%} \\
Codex       & GLM-5.3F   & \cellcolor{codexgray!15} 50.6\% & \cellcolor{codexgray!20} 58.1\% & \cellcolor{codexgray!6} 18.1\% & \cellcolor{codexgray!3} 9.4\% & \cellcolor{codexgray!9} 23.8\% & \cellcolor{codexgray!11} 32.0\% \\
\midrule
Claude Code & DS-Vision  & \cellcolor{ccorange!30} 80.0\% & \cellcolor{ccorange!32} 83.1\% & \cellcolor{ccorange!11} 28.1\% & \cellcolor{ccorange!18} 46.2\% & \cellcolor{ccorange!8} 22.5\% & \cellcolor{ccorange!20} 52.0\% \\
Hermes      & DS-Vision  & \cellcolor{aawblue!29} 76.9\% & \cellcolor{aawblue!35} 91.2\% & \cellcolor{aawblue!12} 30.6\% & \cellcolor{aawblue!15} 39.4\% & \cellcolor{aawblue!10} 27.5\% & \cellcolor{aawblue!21} 53.1\% \\
Codex       & DS-Vision  & \cellcolor{codexgray!33} \textbf{86.9\%} & \cellcolor{codexgray!36} \textbf{94.4\%} & \cellcolor{codexgray!13} \textbf{33.1\%} & \cellcolor{codexgray!22} \textbf{57.5\%} & \cellcolor{codexgray!13} \textbf{34.4\%} & \cellcolor{codexgray!24} \textbf{61.3\%} \\
\midrule
Claude Code & DS-V4.1    & \cellcolor{ccorange!29} 76.2\% & \cellcolor{ccorange!34} 88.1\% & \cellcolor{ccorange!12} 30.6\% & \cellcolor{ccorange!10} 26.2\% & \cellcolor{ccorange!7} 20.0\% & \cellcolor{ccorange!18} 48.2\% \\
Hermes      & DS-V4.1    & \cellcolor{aawblue!32} \textbf{83.8\%} & \cellcolor{aawblue!36} \textbf{92.5\%} & \cellcolor{aawblue!14} \textbf{36.9\%} & \cellcolor{aawblue!20} \textbf{51.2\%} & \cellcolor{aawblue!11} \textbf{28.1\%} & \cellcolor{aawblue!23} \textbf{58.5\%} \\
Codex       & DS-V4.1    & \cellcolor{codexgray!31} 81.9\% & \cellcolor{codexgray!35} 91.9\% & \cellcolor{codexgray!12} 31.2\% & \cellcolor{codexgray!18} 45.6\% & \cellcolor{codexgray!10} 25.6\% & \cellcolor{codexgray!21} 55.2\% \\
\bottomrule
\end{tabular*}
\end{table*}

\subsection{AppWorld Evaluation Results} \label{app:appworld_details}

Table~\ref{tab:appworld-matrix} provides Pass@1 (\%) results on the complete 585-task test suite of AppWorld (168 normal and 417 challenge tasks) across production harnesses for Haiku-4.5, Sonnet-5, and DS-V4.1.

\begin{table}[h]
\caption{
  AppWorld Pass@1 (\%) on the complete 585-task test suite (168 normal and 417 challenge tasks) across production harnesses for Haiku-4.5, Sonnet-5, and DS-V4.1.
  Cell color intensity scales continuously with accuracy within each harness family (terracotta for Claude Code, blue for Hermes, and slate for Codex).
  Bold indicates the highest score per split for each model.
}
\label{tab:appworld-matrix}
\centering
\scriptsize
\setlength{\tabcolsep}{7pt}
\renewcommand{\arraystretch}{1.20}
\begin{tabular}{llccc}
\toprule
\textbf{Harness} & \textbf{Model} & \textbf{Normal} (168) & \textbf{Challenge} (417) & \textbf{Overall} (585) \\
\midrule
Claude Code & Haiku-4.5 & \cellcolor{ccorange!18} 60.7\% & \cellcolor{ccorange!10} 43.4\% & \cellcolor{ccorange!13} 48.4\% \\
Hermes      & Haiku-4.5 & \cellcolor{aawblue!18} 60.1\% & \cellcolor{aawblue!11} \textbf{44.6\%} & \cellcolor{aawblue!13} \textbf{49.1\%} \\
Codex       & Haiku-4.5 & \cellcolor{codexgray!20} \textbf{64.9\%} & \cellcolor{codexgray!10} 43.4\% & \cellcolor{codexgray!13} 49.6\% \\
\midrule
Claude Code & Sonnet-5  & \cellcolor{ccorange!23} 70.8\% & \cellcolor{ccorange!17} \textbf{58.0\%} & \cellcolor{ccorange!18} \textbf{61.7\%} \\
Hermes      & Sonnet-5  & \cellcolor{aawblue!24} \textbf{73.2\%} & \cellcolor{aawblue!14} 52.0\% & \cellcolor{aawblue!17} 58.1\% \\
Codex       & Sonnet-5  & \cellcolor{codexgray!18} 61.3\% & \cellcolor{codexgray!12} 48.4\% & \cellcolor{codexgray!14} 52.1\% \\
\midrule
Claude Code & DS-V4.1   & \cellcolor{ccorange!29} \textbf{81.5\%} & \cellcolor{ccorange!19} 59.5\% & \cellcolor{ccorange!22} 65.8\% \\
Hermes      & DS-V4.1   & \cellcolor{aawblue!20} 64.9\% & \cellcolor{aawblue!13} 47.7\% & \cellcolor{aawblue!15} 52.6\% \\
Codex       & DS-V4.1   & \cellcolor{codexgray!27} 78.0\% & \cellcolor{codexgray!24} \textbf{66.9\%} & \cellcolor{codexgray!25} \textbf{70.1\%} \\
\bottomrule
\end{tabular}
\end{table}

\section{Post-Training Details} \label{app:training}

\subsection{Post-Training Infrastructure and Hyperparameters} \label{sec:training_hyperparameters}

Table~\ref{tab:training-hyperparameters} summarizes the training configuration, hyperparameters, and infrastructure environment used for the SWE-Gym reinforcement learning runs reported in Section~\ref{sec:training_results}.

\begin{table}[h]
\caption{Hyperparameter and infrastructure configuration for SWE-Gym post-training on \aaw{}.}
\label{tab:training-hyperparameters}
\centering
\footnotesize
\setlength{\tabcolsep}{8pt}
\renewcommand{\arraystretch}{1.12}
\begin{tabular}{ll}
\toprule
\textbf{Configuration Field} & \textbf{Setting / Value} \\
\midrule
Base Model & Qwen3.5-4B \\
Reinforcement Learning Algorithm & GRPO \\
Harnesses Evaluated & Claude Code, Codex \\
Learning Rate & $1\times 10^{-6}$ \\
Context Window ($\text{ctx}$) & 50{,}000 tokens \\
Prompts per Batch & 4 \\
Rollouts per Prompt & 16 \\
Training Iterations & 73 iterations (Claude Code), 73 iterations (Codex) \\
Training Instances & SWE-Gym (293 software engineering tasks) \\
Evaluation Suite ($N$) &  SWE-bench Verified (500 instances) \\
Distributed Trainer Integration & Slime framework \\
\bottomrule
\end{tabular}
\end{table}

% \begin{figure*}[h]
% \centering
% \includegraphics[width=\linewidth]{figures/Figure_3_swegym.pdf}
% \caption{SWE-Gym post-training and downstream SWE-bench Verified generalization on \aaw{}.
% (a) GRPO rollout reward across 73 training iterations for a Qwen3.5-4B policy trained under Claude Code (CC) and Codex harnesses, with exponential moving averages ($\alpha{=}0.15$).
% (b) Downstream Pass@1 on SWE-bench Verified comparing base and post-trained checkpoints across harnesses.
% (c) Context efficiency on Claude Code over 42 matched evaluation tasks, where the trained model learns targeted reading, reading fewer lines, avoiding whole-file reads, and rarely hitting the context limit.}
% \label{fig:swegym-training}
% \end{figure*}

\begin{figure*}[t]
\centering
\includegraphics[width=\linewidth]{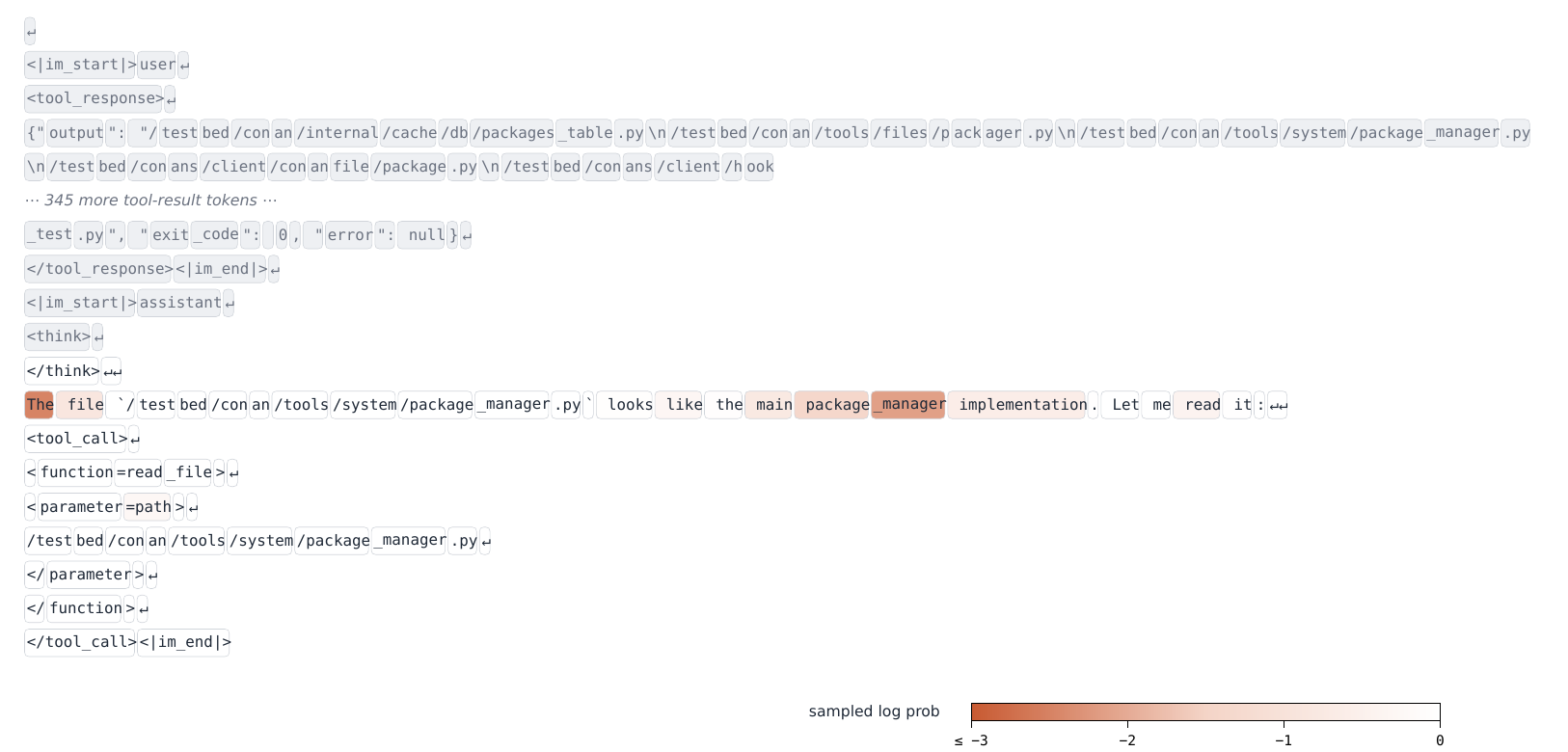}
\caption{Example of tokens and log probabilities captured by the in-container capture proxy, from one model call of a SWE-Gym GRPO training rollout with Qwen3.5-4B (instance \code{conan-io\_\_conan-15931}, training step 20, model call 4 of 36, rollout reward 1.0).
The proxy marks which tokens the model generated and which it did not.
Grey tokens were not generated by the model. They are the prompt tokens added since the previous call, the result of the agent's last tool call and the chat template, and the trainer masks them out of the loss (345 tool-result tokens are elided).
The remaining tokens are the model's completion, each shaded by the log probability the proxy recorded for it at sampling time, and these are the tokens that receive the grader's reward during training.}
\label{fig:rollout-tokens}
\end{figure*}

\subsection{MetaClaw OPD Post-Training Configuration}
\label{app:opd}

Table~\ref{tab:opd-config} summarizes the configuration used for the MetaClaw on-policy distillation
runs reported in Section~\ref{sec:train-metaclaw}. Six runs share this configuration, one per
harness and memory combination.

\begin{figure*}[t]
\centering
\includegraphics[width=0.88\linewidth]{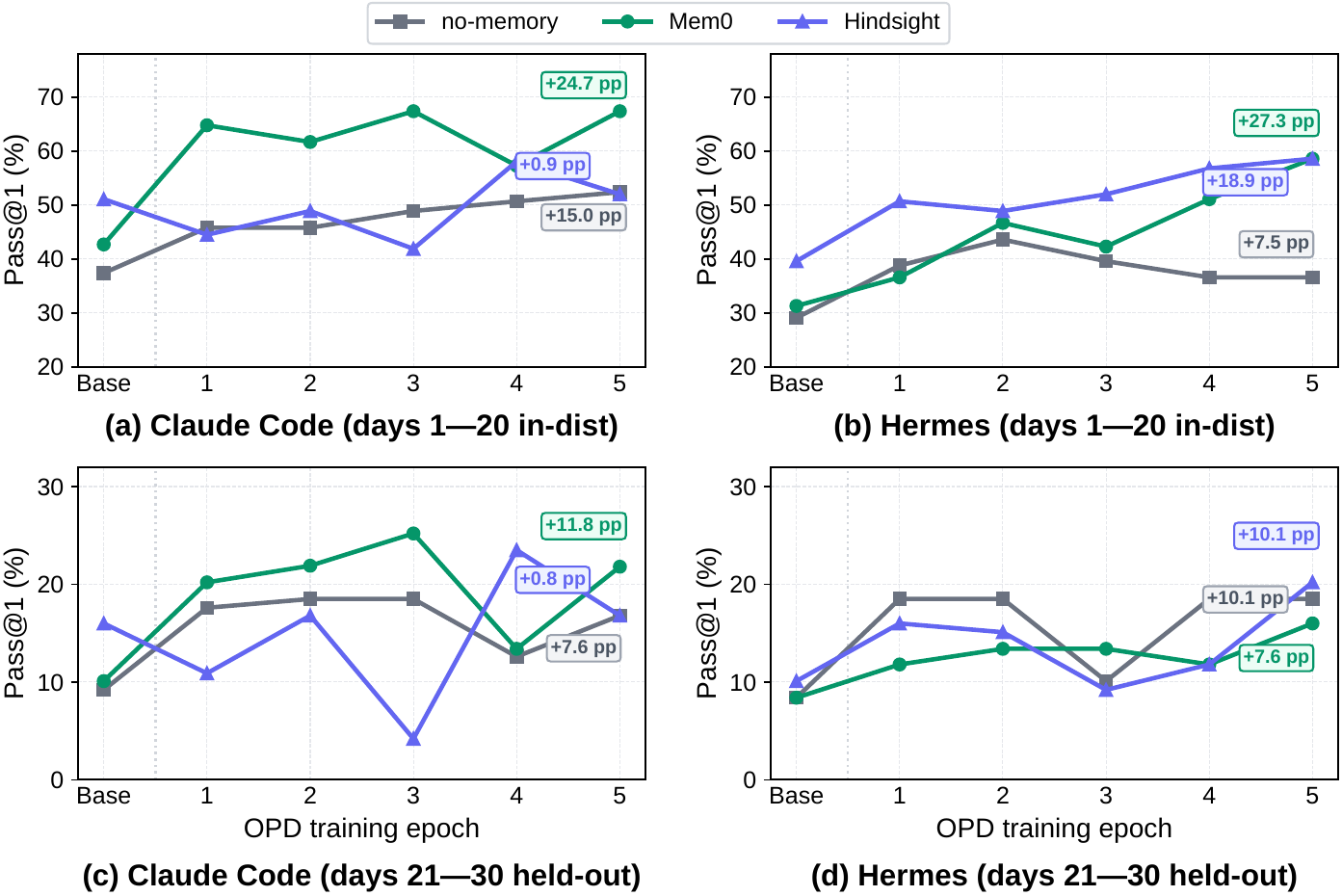}
\caption{Task completion accuracy against OPD training epoch for all six agent and memory configurations.
The top row shows in-distribution performance (days 1--20) for (a) Claude Code and (b) Hermes, and the bottom row shows held-out generalization (days 21--30) for (c) Claude Code and (d) Hermes.
Each point is a separate evaluation over the full 346-round arc, at the base model and at the end of each training epoch.}
\label{fig:opd-epoch}
\end{figure*}

\begin{table}[h]
\centering
\caption{Configuration and infrastructure for MetaClaw OPD post-training on \aaw{}.}
\label{tab:opd-config}
\footnotesize
\begin{tabular}{ll}
\toprule
Configuration Field & Setting / Value \\
\midrule
Student Model & Qwen3.5-4B \\
Teacher Model & Qwen3.5-35B-A3B (MoE), served via SGLang, 65,536-token context \\
Training Algorithm & On-policy distillation via advantage shaping \\
OPD KL Coefficient ($\beta$) & 1.0 \\
Auxiliary Loss Terms & Reference KL 0.001 (\texttt{low\_var\_kl}); entropy coefficient 0.0 \\
Learning Rate & $1 \times 10^{-6}$, constant, no warmup \\
Harnesses Evaluated & Claude Code, Hermes \\
Memory Systems & None, Mem0, Hindsight \\
Prompts per Batch & 1 \\
Rollouts per Prompt & 1 \\
Training Epochs & 5 \\
Max Prompt / Response Tokens & 32,000 / 16,000 \\
Training Instances & MetaClaw (227 rounds over 20 simulated days) \\
Evaluation Suite ($N$) & 346 rounds, split days 1--20 / 21--30 \\
Distributed Trainer Integration & Slime framework \\
\bottomrule
\end{tabular}
\end{table}

\begin{figure}[t]
\centering
\includegraphics[width=\linewidth]{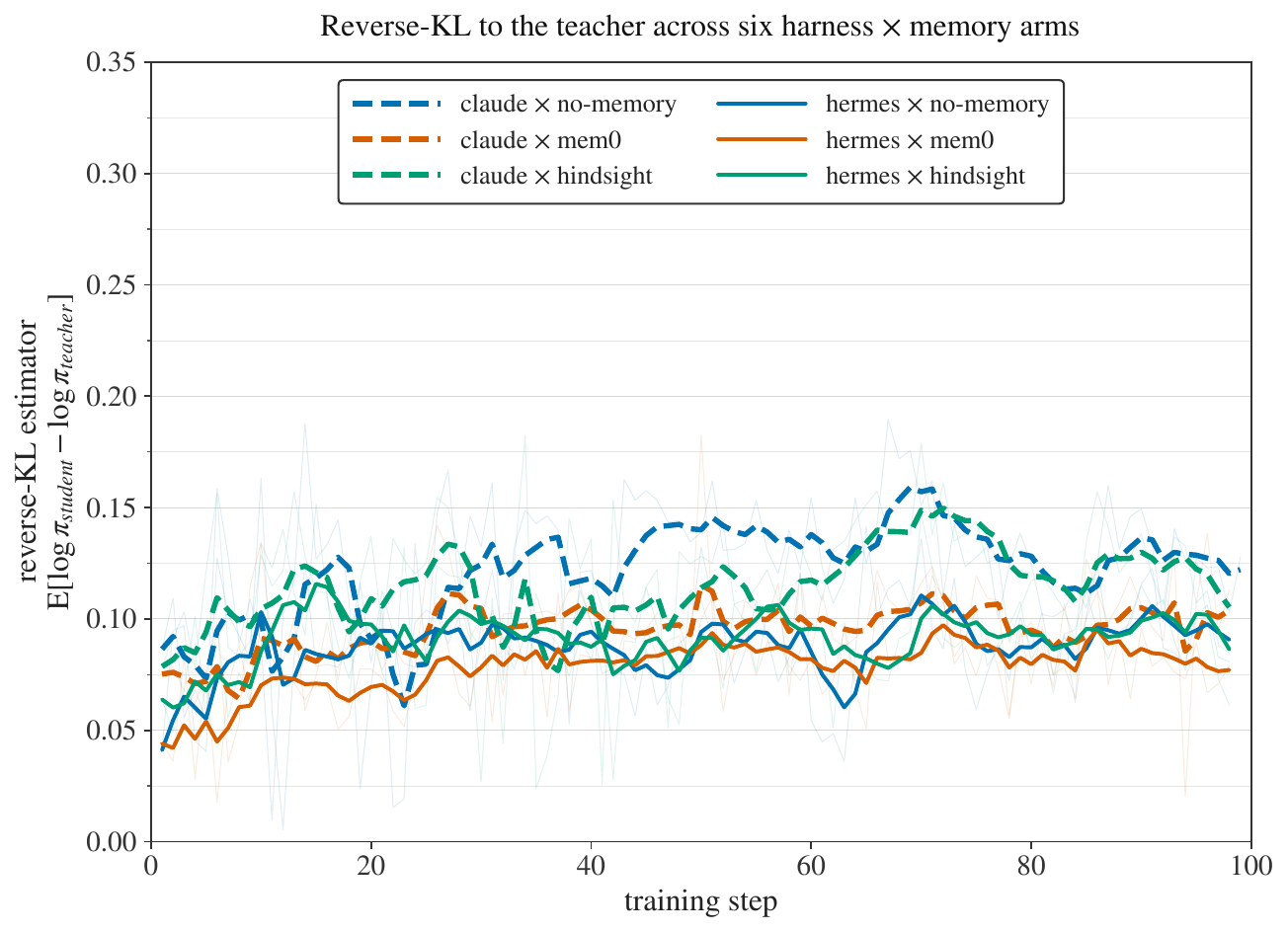}
\caption{Reverse KL between student and teacher logged across training for all six agent and memory
configurations, captured through unmodified harnesses by the in-container proxy. Raw per-step values
are shown under an exponential moving average.}
\label{fig:opd-revkl}
\end{figure}

\noindent \textbf{Objective.} For each sampled response token the trainer forms the log-ratio between student and teacher
log-probabilities, whose expectation under the student's own samples is the reverse KL
$\mathrm{KL}(\pi_{\text{student}} \,\|\, \pi_{\text{teacher}})$, and subtracts it from the advantage:
$\hat{A}_i = A_i - \beta\,(\log \pi_{\text{student}}(y_i) - \log \pi_{\text{teacher}}(y_i))$ with
$\beta = 1.0$. One sample per prompt gives a GRPO group of size one, and no task reward is attached,
so $A_i$ is zero at every logged step and the advantage is $-\beta$ times the log-ratio.
Figure~\ref{fig:opd-revkl} plots its per-step mean for each of the six runs, a sampled estimate of the
reverse KL; the smoothed values remain between 0.04 and 0.16 throughout.

\noindent \textbf{Privileged teacher context.} The teacher is conditioned on a per-round context file the
student never receives, made of 346 JSONL records, one per benchmark round, keyed by day and round
identifier, covering all 30 simulated days (224 \texttt{file\_check} rounds and 122
\texttt{multi\_choice}). Each record carries the round prompt and a privileged-context string, and in
all 346 records that string is the benchmark's own corrective feedback for the round, the text an
agent would otherwise receive only after answering incorrectly. Strings run from 64 to 1,061
characters, median 222.

\subsection{Within-Day Partition}
\label{app:tailcut}

The partition reported in Section~\ref{sec:train-metaclaw} trains on days 1--20 and evaluates on days
21--30. To check the result under a partition in which every evaluation round has training rounds from
the same day, we repeated post-training on a second split of the identical benchmark, in which the model
trains on the first two-thirds of each day's rounds, rounded up, and is evaluated on each day's final
third, holding out 104 of the 346 rounds. The round-level assignment is fixed in advance and shipped
with the scenario.

\begin{figure}[t]
\centering
\includegraphics[width=0.7\linewidth]{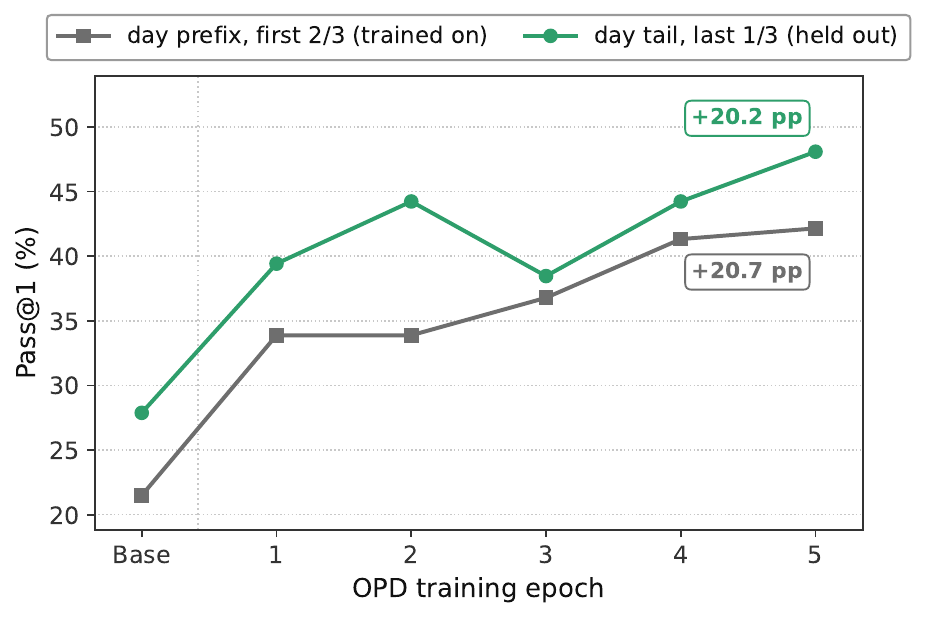}
\caption{Task completion accuracy against OPD training epoch under the within-day partition, Hermes
paired with Mem0. The model trains on the first two-thirds of each day's rounds (242 rounds, gray) and
is evaluated on each day's final third (104 rounds, green).}
\label{fig:opd-tailcut}
\end{figure}

The configuration matches Table~\ref{tab:opd-config} with three changes, as training covers all 30
simulated days rather than days 1--20, the arm is Hermes paired with Mem0, and evaluation applies the
within-day split. Six checkpoints were evaluated, the untrained base and epochs 1--5, each over all 346
rounds (242 in the training slice, 104 held out). Accuracy on the training slice rises from 21.5\% to
42.1\% and on the held-out tail from 27.9\% to 48.1\% (Figure~\ref{fig:opd-tailcut}).

\section{Memory-Write Analysis Controls and Additional Measurements}
\label{app:memwrite}

This appendix supports the qualitative analysis in Section~\ref{sec:qual-memwrite}. All
measurements are over the two full stores of the Hermes${\times}$Hindsight configuration, 5{,}027
facts for the untrained extractor and 3{,}617 for the distilled one at epoch 5. Both arms run with
the memory tools forced on, write through the same Hindsight daemon, and share a byte-identical
container image, an identical launch command, and an identical extraction prompt; the only
environment keys that differ are the served model and the two upstream URLs that follow it.

\noindent \textbf{Where the schema comes from.} Each stored fact follows a pipe-delimited template,

\begin{center}
\texttt{statement | When: date | Involving: who | To purpose}
\end{center}

\noindent and the template is the memory system's, not something the model invents. We set no
retain mission or extraction prompt of our own, so extraction runs on the backend's default. Two
observations establish that the structure is imposed rather than emergent. The field order is
canonical in $100\%$ of facts carrying both a date and an entity, in both arms, and the untrained
$4$B extractor already emits a \texttt{When} field on $99.2\%$ of its facts. What training moves is
the one field that is effectively discretionary. Field presence for untrained against distilled is
$99.2\%$ to $99.9\%$ for \texttt{When}, $83.6\%$ to $81.8\%$ for \texttt{Involving}, and $1.1\%$ to
$19.6\%$ for the purpose clause. The \texttt{Involving} field is the useful control here, neither saturated nor absent, and it does not move, so the purpose-clause gain is not a general
increase in template compliance.

\begin{figure}[t]
\centering
\includegraphics[width=\linewidth]{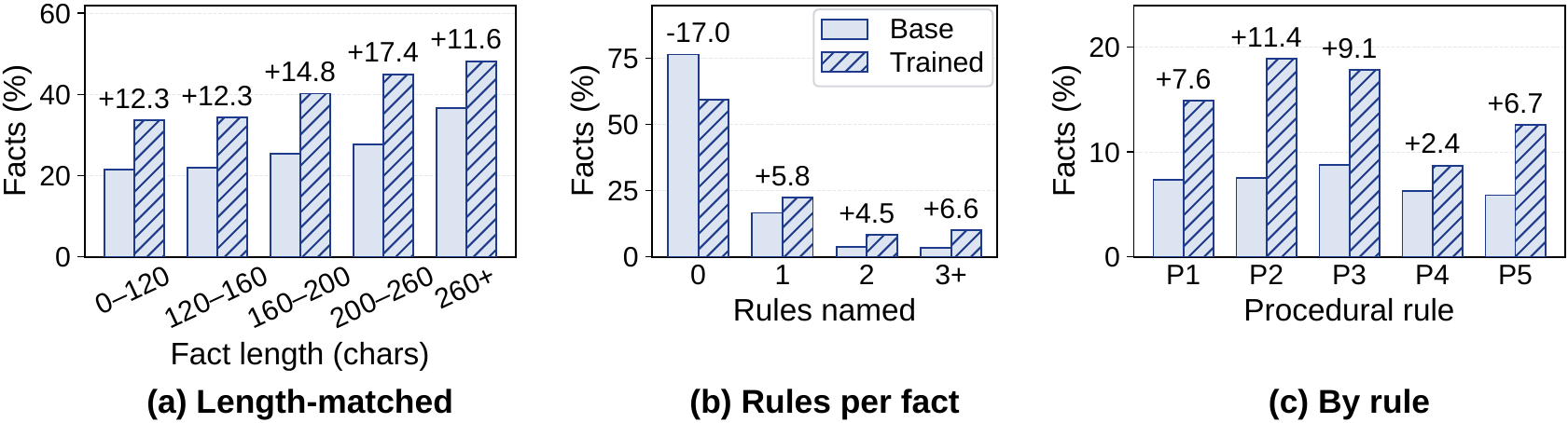}
\caption{The procedural-rule gain is structural rather than a by-product of longer facts, Hermes agent with the Hindsight backend, both arms with the memory tools forced on.
(a) Facts naming at least one procedural rule within fact-length strata; the trained store leads in every stratum.
(b) The distribution over the number of distinct rules a fact names shifts rather than scaling, with facts naming no rule falling $17.0$ pp.
(c) Per-rule incidence; all five rise, least for the backup rule (P4).}
\label{fig:memwrite-mechanism}
\end{figure}

\noindent \textbf{Length controls.} Two measurements establish that the procedural-rule gain is not an
artifact of trained facts being 42 characters longer. Stratifying facts by total length, the gap holds
inside every stratum, at 12.3, 12.3, 14.8, 17.4 and 11.6 pp from shortest to longest, so there is no
length band in which the two stores coincide (Figure~\ref{fig:memwrite-mechanism}a). Normalising for
length directly, distinct rules named per 100 characters rises from 0.240 to 0.378.

\noindent \textbf{Distribution and per-rule detail.} The distribution over the number of distinct
procedural rules a fact names shifts rather than scaling
(Figure~\ref{fig:memwrite-mechanism}b). Facts naming none fall $17.0$ pp, from $76.4\%$ to
$59.4\%$, while facts naming three or more rise from $3.4\%$ to $10.0\%$. All five procedural rules rise (Figure~\ref{fig:memwrite-mechanism}c), by factors of 2.04, 2.52, 2.04, 1.38 and 2.14, the smallest being the backup rule.

\begin{figure}[t]
\centering
\includegraphics[width=\linewidth]{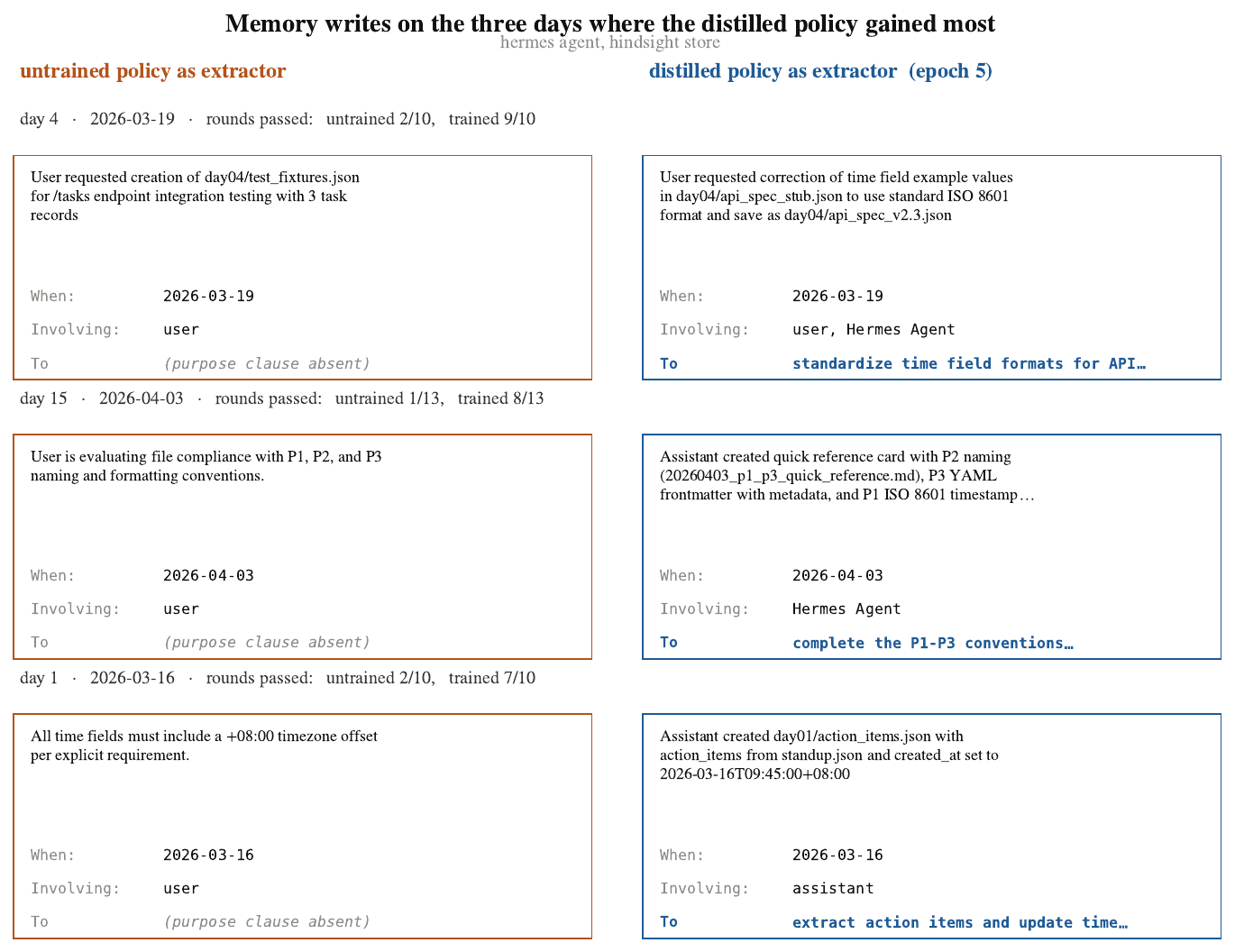}
\caption{Verbatim memory writes from the two arms on three of the days where the distilled model's
gain in rounds passed is largest, Hermes agent with the Hindsight backend. Within each day, each cell
is the most completely filled record that store wrote, scored identically for both arms. The distilled
records name the artifact produced, attribute it to the agent, and state a purpose; the untrained
records restate the instruction or describe the user. The round tally counts the rounds passed on that
day and is not tied to the round that produced the record. Fact text
is verbatim but not de-duplicated.}
\label{fig:memwrite-examples}
\end{figure}

\noindent \textbf{Verbatim examples.} Figure~\ref{fig:memwrite-examples} shows matched records from the two
stores. Neither the days nor the records are chosen by inspection. The three days are the two on which the distilled model's gain in rounds passed is largest (+7) and
the earliest of the three tied next (+5); the distilled model passes strictly more rounds on 23 of the 30 days.
Within each day, each cell is the most completely filled record that store wrote, scored identically
for both arms by schema fields present and then by rules named, so neither arm is shown at a
random draw. On all three days the untrained store contains no fact with a purpose clause at all,
$0$ of $90$, $0$ of $189$ and $0$ of $161$, which is why its ceiling in the figure is a three-field
record. One pair is worth reading closely because both of its records name the same three
procedural rules, so the difference between them is not rule count. The untrained extractor writes
``User is evaluating file compliance with P1, P2, and P3 naming and formatting conventions'' with no
purpose field, while the distilled one writes ``Assistant created quick reference card with P2
naming, P3 YAML frontmatter with metadata, and P1 ISO 8601 timestamp format'', attributed to the
agent and carrying the purpose ``to complete the P1-P3 conventions reference card''. The round tally shown in the figure counts the rounds passed on that day and is not tied to the round that produced the record.

\noindent \textbf{Multi-entity records.} Facts naming more than one entity in the \texttt{Involving} field
rise from $0.6\%$ to $8.5\%$, consistent with the subject shift reported in the main text.

\end{document}